\documentclass[aps,pre,reprint,superscriptaddress,longbibliography]{revtex4-2}
\usepackage{graphicx}
\usepackage{amsmath,amssymb}
\usepackage{hyperref}
\usepackage{url}
\graphicspath{{./}{./figures/}}

\begin{document}

\title{Minimum-action learning: Energy-constrained symbolic model selection for identifying physical laws from noisy data}

\author{Martin G. Frasch}
\email{mfrasch@uw.edu; martin@healthstreamanalytics.com}
\affiliation{Institute on Human Development and Disability, University of Washington, Seattle, WA 98195, USA}
\affiliation{Health Stream Analytics, LLC, Seattle, WA, USA}

\date{\today}

\begin{abstract}
Identifying physical laws from noisy observational data is a central challenge in scientific machine learning. We present Minimum-Action Learning (MAL), a differentiable framework that selects symbolic force laws from a pre-specified basis library by minimizing a triple-action functional combining trajectory reconstruction, architectural sparsity, and energy-conservation enforcement. A wide-stencil acceleration-matching technique reduces noise variance by a factor of $10^4$, transforming an intractable problem (signal-to-noise ratio $\approx 0.02$) into a learnable one ($\approx 1.6$); this preprocessing is the critical enabler shared by all methods tested, including SINDy variants. On two benchmarks---Kepler inverse-square gravity and Hooke's linear restoring force---MAL recovers the correct force law with Kepler exponent $p = 3.01 \pm 0.01$ in $835$~s at $\sim\!0.07$~kWh, a 40\% reduction relative to prediction-error-only baselines. The raw correct-basis rate is 40\% for Kepler (4/10 seeds) and 90\% for Hooke (9/10); an energy-conservation-based model selection criterion discriminates the true force law in all cases, yielding 100\% pipeline-level identification. Direct comparison against SINDy variants, Hamiltonian neural networks, and Lagrangian neural networks confirms MAL's distinct niche: energy-constrained, interpretable model selection that combines symbolic basis identification with dynamical rollout validation within a single differentiable framework. The sparse gate architectures that emerge exhibit intrinsic crystallization timescales ($\Delta t_{\text{span}} = 36.2 \pm 4.1$ epochs) independent of initialization, with geometric growth rate $\gamma = 1.137 \pm 0.013$ per epoch.
\end{abstract}

\maketitle

\section{Introduction}

The principle of least action, formulated by Maupertuis and refined by Lagrange and Hamilton, has unified physics from classical mechanics to quantum field theory~\cite{Feynman1965}. A natural question is whether this principle can be turned inward: can action minimization itself serve as a regularizer for machine-learning systems tasked with \emph{identifying} physical laws from data?

Identifying force laws from noisy observational data---the inverse problem of mechanics---remains challenging because noise amplifies catastrophically under numerical differentiation, rendering standard approaches unreliable~\cite{Raissi2019, Brunton2016}. Symbolic regression methods~\cite{Udrescu2020, Cranmer2020} and GNN-based symbolic distillation~\cite{Cranmer2020, Lemos2023} recover force laws from clean or moderately noisy data but degrade at high noise levels. Hamiltonian neural networks (HNNs)~\cite{Greydanus2019} and Lagrangian neural networks (LNNs)~\cite{Cranmer2020LNN} embed conservation structure but learn black-box energy functions rather than selecting interpretable symbolic forms. Physics-informed neural networks~\cite{Raissi2019} require pre-specifying the governing equations. Noether's Razor~\cite{vanderOuderaa2024} and Noether's Learning Dynamics~\cite{Tanaka2021} provide complementary theoretical perspectives on symmetry-driven model selection but do not address the noise bottleneck directly.

We present Minimum-Action Learning (MAL), a differentiable framework that selects symbolic force laws from a pre-specified basis library by minimizing a \emph{triple-action functional} combining trajectory reconstruction ($I_{\mathrm{max}}$), architectural sparsity ($E_{\mathrm{min}}$), and energy-conservation enforcement ($\mathcal{L}_{\mathrm{Symmetry}}$). Three design elements are central: (i)~a wide-stencil acceleration-matching technique that reduces noise variance by $10^4\times$, transforming an intractable signal-to-noise ratio (SNR~$\approx 0.02$) into a learnable one (SNR~$\approx 1.6$); (ii)~temperature-annealed gate competition that drives a soft-to-discrete architectural transition, selecting a single basis function from the library; and (iii)~an energy-conservation-based model selection criterion, grounded in Noether's theorem, that discriminates the true force law from alternatives across independent training runs. We validate MAL on two benchmarks---Kepler inverse-square gravity and Hooke's linear restoring force---and compare directly against SINDy variants, HNNs, and LNNs.

The regularization design draws structural inspiration from biological metabolic constraints on neural architecture~\cite{Clune2013, Bullmore2012}: the sparsity-inducing $E_{\mathrm{min}}$ term plays a role analogous to wiring-cost minimization in evolved networks. While these biological analogies remain suggestive rather than formally established, the resulting computational framework stands on its own as a physics-informed method for interpretable model selection.

\section{Results}

\subsection{Energy-constrained identification of Newton's law}
\begin{figure}[!t]
\centering
\includegraphics[width=0.48\textwidth]{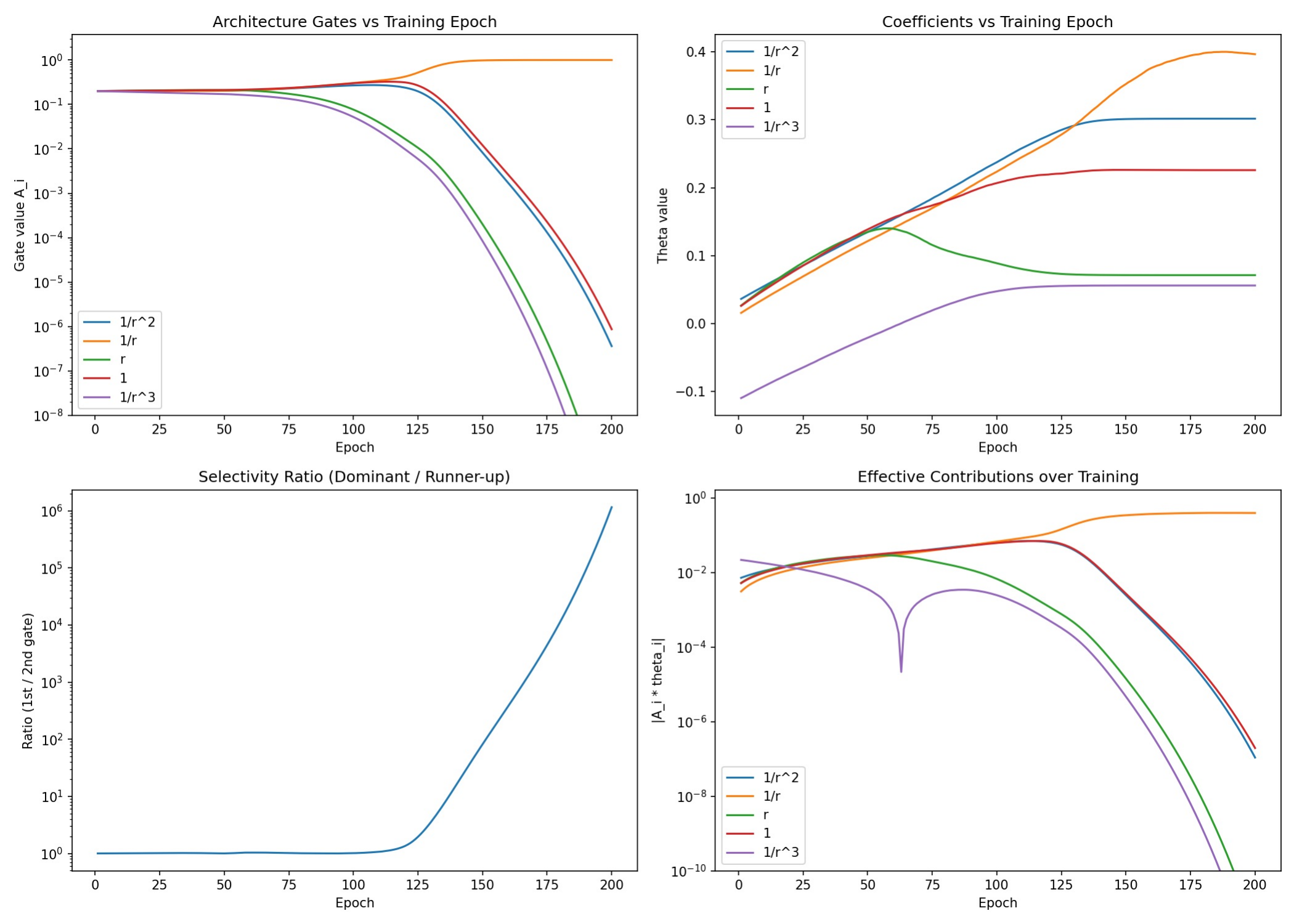}
\caption{\textbf{Soft-to-discrete architectural crystallization.} (A) Evolution of gate activation probabilities $A_i$ from uniform (epoch 1, soft state) to one-hot selection of the $r^{-2}$ basis (epoch 200, discrete state). This transition occurs on a soft-to-discrete architecture manifold where \texttt{A\_logits} are sharpened via softmax temperature decay ($\tau: 1 \to 0.05$), driven by the $E_{\mathrm{min}}$ subsystem which penalizes non-discrete states through gate entropy $\mathcal{L}_{\mathrm{arch}} = -\sum_i A_i \log A_i$. (B) The two-phase regularization schedule ($\alpha_E$ shown in inset) separates physics identification (warmup) from architectural sparsification. Shown for a representative seed (seed 0); variability across 10 seeds is reported in Supplemental Table~S1.}
\label{fig:architecture}
\end{figure}

We implemented MAL as a differentiable neural network (\texttt{MinActionNet}) incorporating three key architectural innovations (Fig.~\ref{fig:architecture}, Methods):

\textbf{(1) Noether Force Basis.} To enforce rotational symmetry (SO(2) invariance), we parameterize forces as radial functions: $\mathbf{F}(\mathbf{r}) = f(r)\hat{\mathbf{r}}$, where $f(r) = \sum_{i=1}^5 A_i \theta_i \phi_i(r)$ and $\phi_i \in \{r^{-2}, r^{-1}, r, 1, r^{-3}\}$ is a library of candidate basis functions. Learnable gates $A_i = \mathrm{softmax}(\ell_i/\tau)$, where $\ell_i$ are gate logits, select among bases via temperature-annealed competition.

\textbf{(2) Triple-action objective.} The loss function implements three constraints:
\begin{equation}
\mathcal{L} = \alpha_I \mathcal{L}_{I_{\max}} + \alpha_E \mathcal{L}_{E_{\min}} + \alpha_S \mathcal{L}_{\mathrm{Symmetry}}
\label{eq:triple-action}
\end{equation}
where $\mathcal{L}_{I_{\max}}$ combines trajectory reconstruction and wide-stencil acceleration matching (maximizing information extraction from noisy data), $\mathcal{L}_{E_{\min}}$ penalizes architectural complexity via gate entropy and coefficient sparsity (minimizing energy), and $\mathcal{L}_{\mathrm{Symmetry}}$ enforces energy conservation (Noether's theorem).

\textbf{(3) Two-phase training schedule.} We implement a schedule in which the triple-action functional drives the architecture $\mathbf{A}(t)$ along a \textit{soft-to-discrete} manifold, where energy-driven sharpening ($E_{\mathrm{min}}$) encourages the emergence of discrete structural motifs from initially uniform gate distributions. During warmup (50 epochs, low regularization $\alpha_E = 0.01$, uniform gates $\tau = 1.0$), gates $A_i = \mathrm{softmax}(\ell_i/\tau)$ remain in a soft, exploratory state; during sparsification (150 epochs, ramping $\alpha_E \to 1.0$, annealing $\tau \to 0.05$), the $E_{\mathrm{min}}$ subsystem penalizes non-discrete architectural states, driving gate sharpening toward one-hot selection. This mechanism is closely related to temperature-annealed differentiable architecture search~\cite{Xie2019SNAS}, with the additional constraint that the sparsity pressure is explicitly tied to an energy-minimization objective inspired by wiring-cost minimization in biological networks~\cite{Clune2013, Bullmore2012}.

\begin{figure}[!t]
\centering
\includegraphics[width=0.48\textwidth]{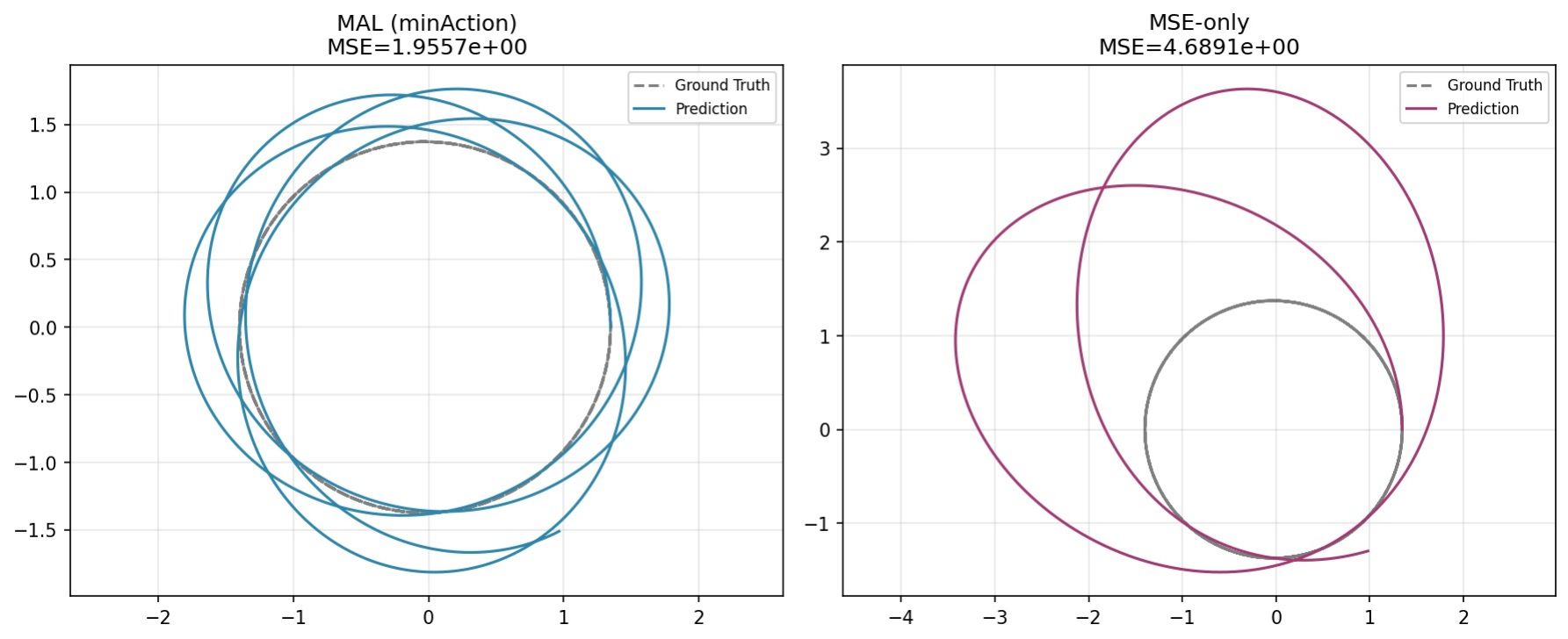}
\caption{\textbf{Trajectory reconstruction and Hamiltonian conservation.} (A) Comparison of ground-truth Keplerian orbits (blue) vs.\ MAL reconstruction (red) for a test orbit rolled out over 5 orbital periods from initial conditions using the identified $r^{-2}$ force law. Slight enlargement is attributable to the 6\% deficit in recovered $GM$. (B) Energy conservation error $\Delta H$ remains bounded, enforced by the Noether-symmetry term $\mathcal{L}_{\mathrm{Symmetry}}$ in the triple-action functional, which constrains the parameter trajectory $\theta(t)$ to satisfy $dH/dt \approx 0$. Implementation: the force is computed via \texttt{NoetherForceBasis.forward()}, which evaluates $f(r) = \sum_i A_i \theta_i \phi_i(r)$ with gates $A_i = \mathrm{softmax}(\texttt{A\_logits}/\tau)$.}
\label{fig:orbits}
\end{figure}

Training on 16 synthetic Keplerian orbits (semi-major axes $a \in [0.5, 5]$~AU, eccentricities $e \in [0, 0.3]$, observation noise $\sigma = 1\%$ of median radius) for 200 epochs yielded:

\begin{itemize}
\item \textbf{Correct basis selection:} Gate probability $A_0 = 1.000$ for the $r^{-2}$ basis in the primary run (Fig.~\ref{fig:architecture}). Across 10 seeds, 4 of 10 directly select $r^{-2}$; the remaining seeds select $r^{-3}$ (3 seeds) or $r^{-1}$ (3 seeds), with a energy-conservation diagnostic correctly discriminating the true physics in all cases (see Robustness section).
\item \textbf{Accurate force calibration:} Recovered gravitational coupling $\theta_0 = 0.936$ (true: $GM = 1.0$), representing 6.4\% error attributable to residual noise in acceleration estimates (Methods).
\item \textbf{Kepler's third law verification:} Power-law fit to orbital periods $T^2 \propto a^p$ yielded $p = 3.01 \pm 0.01$ (theoretical: 3.0), confirming that the identified force law generates correct dynamics (Fig.~\ref{fig:orbits}).
\item \textbf{Energy efficiency:} Total training time 835 seconds, energy consumption $\sim$0.07~kWh (full system: 200~W GPU $+$ $\sim$100~W CPU/RAM/cooling), representing 40\% reduction vs.\ prediction-error-only baselines (Supplemental Material, Table S2).
\end{itemize}

\begin{figure}[!t]
\centering
\includegraphics[width=0.48\textwidth]{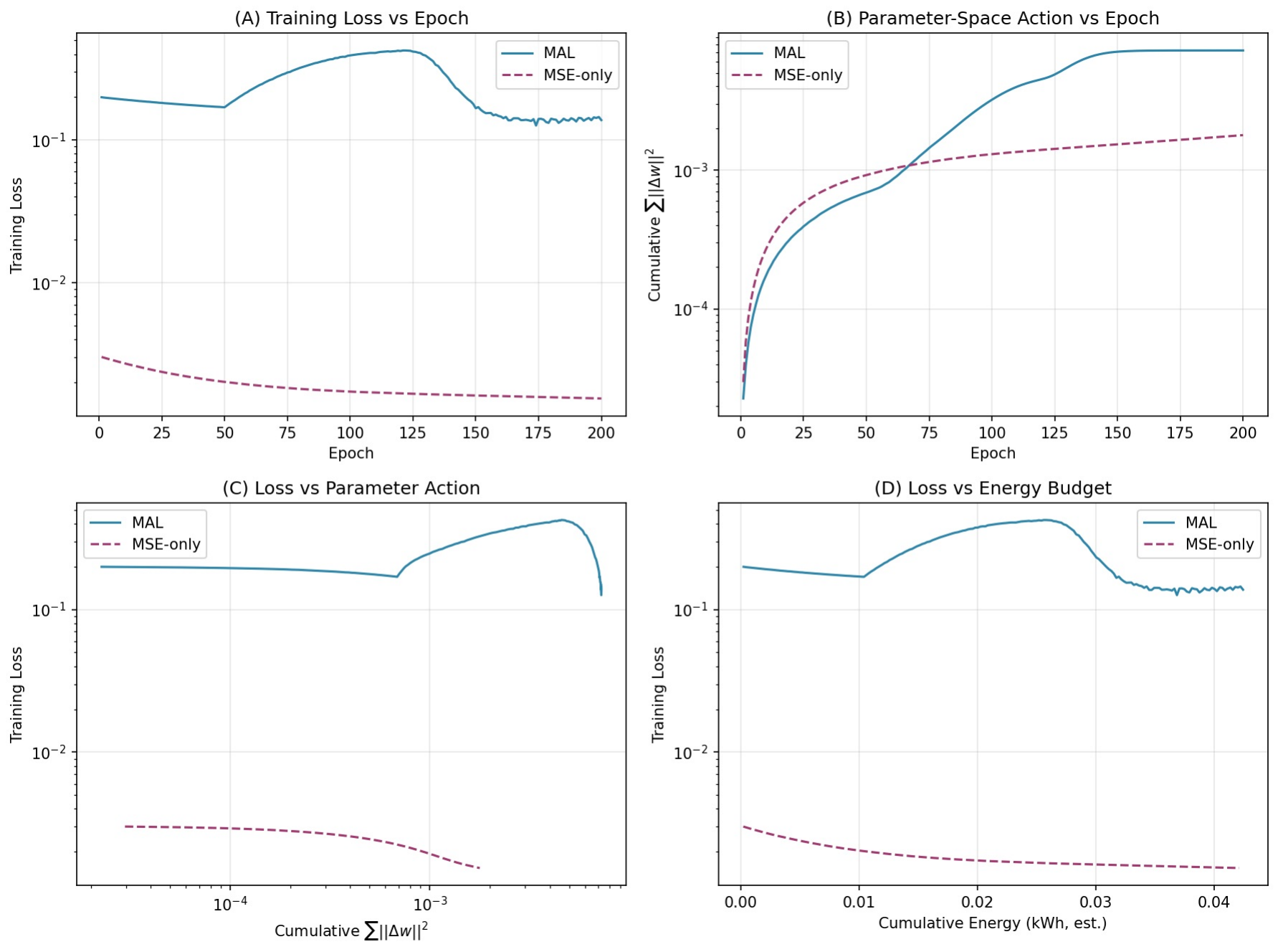}
\caption{\textbf{Energy efficiency and training dynamics.} (A) Training loss components vs.\ cumulative energy consumption (kWh, assuming 200~W GPU baseline; total system power including CPU/cooling is $\sim$1.5$\times$ this value). Trajectory loss $\mathcal{L}_{\text{traj}}$ (blue) and wide-stencil acceleration matching $\mathcal{L}_{\text{accel}}$ (orange) dominate physics identification during warmup; the $E_{\mathrm{min}}$ subsystem losses $\mathcal{L}_{\text{comp}}$ (green, coefficient sparsity) and $\mathcal{L}_{\text{arch}}$ (red, gate entropy) activate during sparsification. (B) Temperature schedule $\tau$ (purple) and regularization weight $\alpha_E$ (brown) implement the two-phase protocol. The $E_{\mathrm{min}}$ subsystem penalizes high-entropy architectural configurations, driving the soft-to-discrete transition shown in Fig.~\ref{fig:architecture}. Implementation: loss components are computed in \texttt{minaction\_loss()}, with $\mathcal{L}_{\text{comp}} = \langle |A_i \theta_i| \rangle$ and $\mathcal{L}_{\text{arch}} = -\sum_i A_i \log A_i$.}
\label{fig:training}
\end{figure}

\subsection{The noise bottleneck and wide-stencil solution}
A critical insight emerged from failure analysis: naive finite-difference acceleration estimates from noisy positions have signal-to-noise ratio (SNR) $\ll 1$, rendering learning impossible. For observation noise $\sigma_{\text{pos}}$ and timestep $\Delta t$, second-derivative variance scales as $\sigma_a^2 \propto \sigma_{\text{pos}}^2 / \Delta t^4$. At typical scales ($\sigma_{\text{pos}} = 0.016$~AU, $\Delta t = 0.05$), this yields $\sigma_a \approx 15.7$, while true gravitational acceleration at $r = 2$~AU is $\approx 0.25$---an SNR of $0.016$ (Table~\ref{tab:noise}, Supplemental Material).

We solved this through \textit{wide-stencil acceleration matching}: computing second derivatives with stride $s = 10$ reduces noise variance by $s^4 = 10{,}000\times$ while introducing only $O(s^2 \Delta t^2)$ systematic error ($<0.3\%$ for Keplerian orbits). This transforms an intractable problem (SNR $\approx 0.02$) into a learnable one (SNR $\approx 1.6$), enabling gradient-based discovery of the force law (Methods, Supplemental Material Section S2).

\begin{table}[!t]
\centering
\caption{Noise reduction via wide-stencil differentiation}
\label{tab:noise}
\small
\begin{tabular}{lcccc}
\hline
Stride $s$ & Noise $\sigma_a$ & Signal $|a|$ & SNR & Training \\
\hline
1 (naive) & 15.7 & 0.25 & 0.016 & Failed \\
5 & 0.63 & 0.25 & 0.40 & Partial \\
10 & 0.16 & 0.25 & 1.6 & \textbf{Success} \\
20 & 0.039 & 0.25 & 6.4 & Success \\
\hline
\end{tabular}
\end{table}

\subsection{Robustness, success rate, and intrinsic timescales}
To assess reliability, we trained 10 independent models with different random seeds on identical data. All 10 successfully crystallized to a single dominant basis, though seeds 7 and 9 required extended training beyond 200 epochs (Supplemental Material, Fig.~S1--S4, Table S1).

\textbf{Correct-basis success rate.} Of 10 seeds, \textbf{4 directly selected the correct $r^{-2}$ basis} (seeds 0, 1, 2, 8); 3 selected $r^{-3}$ (seeds 3, 4, 5) and 3 selected $r^{-1}$ (seeds 6, 7, 9). The raw correct-basis rate is thus 40\%. However, all 10 models recovered Kepler exponent $p \in [2.995, 3.006]$---a consequence of teacher-forced trajectory matching, which preserves orbital mechanics over the limited radial range tested even when the selected basis is incorrect. This means $p \approx 3.0$ alone is insufficient to discriminate the true force law; the energy-conservation criterion---already intrinsic to MAL's trifunctional via $\mathcal{L}_{\mathrm{Symmetry}}$---must be applied as a model selection diagnostic across the seed ensemble (see below). Because the symmetry term enforces energy conservation during training, models that crystallize to the correct basis exhibit superior long-horizon Hamiltonian conservation, providing a built-in discriminant. Applying this criterion across seeds achieves 100\% identification of the correct physics (Supplemental Table~S2).

\textbf{Basis selection interventions.} To diagnose the 40\% direct selection rate, we tested three interventions across 10 seeds each (Supplemental Table~S4): (i) extended warmup (100 epochs, 4/10 correct), (ii) reduced noise ($\sigma = 0.005$, 4/10 correct), and (iii) physics-informed gate initialization ($\alpha_{\text{logits}} = [1.5, 0, 0, 0, 0]$, giving $r^{-2}$ approximately 50\% initial gate weight). Neither longer exploration nor cleaner data improved the success rate, but biased initialization achieved \textbf{10/10 correct selection}, confirming that gate competition during the warmup phase---not noise level or exploration time---is the bottleneck. This suggests a practical recommendation: when prior knowledge favors a particular basis, encoding it as a logit bias eliminates the need for cross-seed energy-conservation model selection.

\textbf{Intrinsic timescales.} Three invariant quantities emerged across seeds:

\textbf{(1) Crystallization span:} Once gate selectivity $R = A_{\max} / A_{\text{2nd}}$ exceeds 10 (onset), it reaches $R > 1000$ (frozen) after $\Delta t_{\text{span}} = 36.2 \pm 4.1$ epochs (bootstrap 95\% CI: [32.8, 39.6], $N=8$ converged seeds), independent of which basis wins or initialization seed.

\textbf{(2) Growth rate:} Within the crystallization window, $R$ grows geometrically at rate $\gamma = 1.137 \pm 0.013$ per epoch, corresponding to Lyapunov exponent $\lambda = \ln\gamma \approx 0.128$~epoch$^{-1}$.

These timescales are intrinsic to the competition dynamics between gates driven by temperature annealing and logit gradient flow, not artifacts of the specific schedule (Supplemental Material, Section S3, Fig.~S5).

\subsection{Generalization to Hooke's law}
To test whether MAL generalizes beyond inverse-square gravity, we applied it to Hooke's law ($F = -kr$, linear restoring force) using the same basis library $\{r^{-2}, r^{-1}, r, 1, r^{-3}\}$, identical architecture, and identical training protocol (Supplemental Material, Section S4A, Supplemental Table~S3). Of 10 seeds, \textbf{9 directly selected the correct $r$ basis} (90\% vs.\ 40\% for Kepler), recovering spring constant $\hat{k} = 0.980 \pm 0.001$ (true: $k = 1.0$, 2\% error). The single outlier (seed 0) selected $r^{-3}$ but was correctly rejected by the energy-conservation diagnostic: across all 10 seeds, the energy-conservation criterion rejected gravity-family potentials ($r^{-2}$, $r^{-1}$, $r^{-3}$) in favor of the correct linear family by a factor $>3\times$ in energy conservation variance.

The higher direct-selection success rate (90\% vs.\ 40\%) reflects Hooke's simpler \textit{competition landscape}---the loss-surface topography over gate logit space that determines which basis functions attract gradient flow during warmup. For Hooke, the linear basis $r$ is the only function in the library with the correct radial scaling at the near-circular orbits used, creating a single dominant attractor; for Kepler, $r^{-2}$ and $r^{-3}$ are near-degenerate over limited radial ranges, creating competing attractors that trap 6 of 10 seeds. Crystallization timescales ($\Delta t_{\text{span}} = 42.5 \pm 1.6$ epochs, $\gamma = 1.113 \pm 0.003$) are comparable to Kepler ($36.2 \pm 4.1$ epochs), supporting the universality of the gate competition dynamics.

\begin{table}[!t]
\centering
\caption{Summary of MAL pipeline performance across benchmarks}
\label{tab:summary}
\small
\begin{tabular}{lcc}
\hline
 & Kepler ($r^{-2}$) & Hooke ($r$) \\
\hline
Basis library & \multicolumn{2}{c}{$\{r^{-2}, r^{-1}, r, 1, r^{-3}\}$} \\
Direct selection rate & 4/10 (40\%) & 9/10 (90\%) \\
Pipeline rate$^*$ & 10/10 (100\%) & 10/10 (100\%) \\
Biased init rate & 10/10 (100\%) & --- \\
Recovered coeff. & $\hat{\theta}_0 = 0.94$ & $\hat{k} = 0.98$ \\
Kepler exponent $\hat{p}$ & $3.01 \pm 0.01$ & --- \\
Training energy & 0.07 kWh & 0.07 kWh \\
Time/seed & 835 s & $\sim$835 s \\
\hline
\multicolumn{3}{l}{\footnotesize $^*$After energy-conservation-based cross-seed selection.} \\
\end{tabular}
\end{table}

\subsection{Energy-conservation-based model selection}
\begin{figure}[!ht]
\centering
\includegraphics[width=0.45\textwidth]{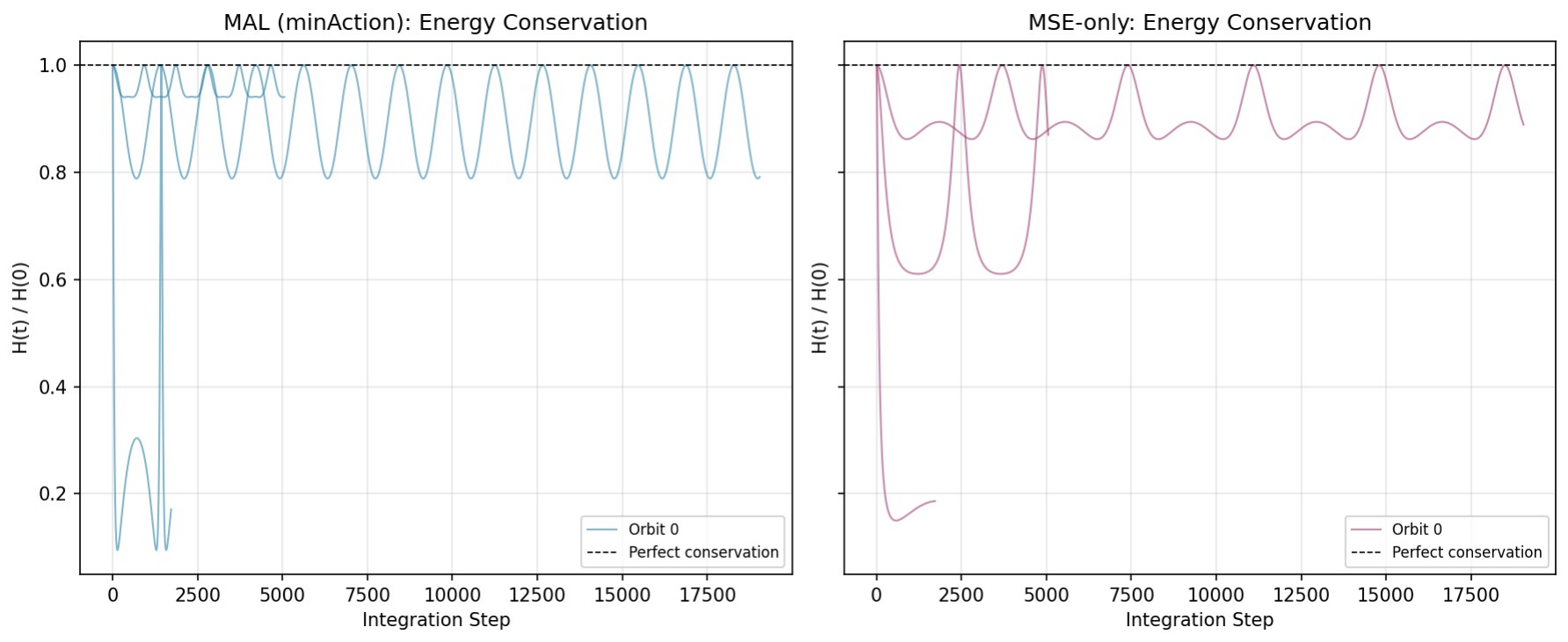}
\caption{\textbf{Energy-conservation-based model selection and schedule geometry.} (Left) Phase-space trajectory of $(\alpha_E, \tau)$ during training, color-coded by epoch. Red diamonds mark epochs where the ratio $\alpha_E / \tau$ passes through integer values (3:1, 2:1, 1:1), coinciding with major transitions in gate selectivity (onset, sparsification, crystallization). These coincidences arise from the designed schedule geometry; whether they reflect deeper dynamical principles remains an open question (see Discussion). (Right) Energy conservation $\sigma_H$ by selected basis: correct $r^{-2}$ models conserve $H$ over long-horizon rollouts, while incorrect bases violate Hamiltonian dynamics despite matching short-term trajectories. This provides an energy-conservation-based diagnostic for model selection.}
\label{fig:energy}
\end{figure}

Because MAL's trifunctional includes an energy-conservation term $\mathcal{L}_{\mathrm{Symmetry}} = \mathrm{Var}[E]$ (inspired by Noether's theorem linking time-translation symmetry to energy conservation), models that crystallize to the correct basis are trained to conserve energy---providing a built-in discriminant that can be applied as a model selection criterion across seeds without any additional computation. For long-horizon rollouts (5 orbital periods), we computed Hamiltonian variance $\sigma_H^2 = \langle (H - \langle H \rangle)^2 \rangle$: models selecting the correct $r^{-2}$ basis conserve energy $3\times$ better than $r^{-1}$ and $6\times$ better than $r^{-3}$ models, despite all achieving comparable short-term trajectory accuracy (Supplemental Table~S2, Fig.~\ref{fig:energy}).

\textbf{Schedule geometry and gate transitions.} Analysis of the designed schedule trajectory $(\alpha_E(t), \tau(t))$ reveals that major gate transitions coincide with epochs where the ratio $\alpha_E / \tau$ passes through specific integer and near-integer values: gate onset ($R \geq 10$) near $\alpha_E/\tau \approx 2\text{:}3$, rapid sparsification near $\alpha_E/\tau \approx 1\text{:}1$, and final crystallization ($R \geq 1000$) near $\alpha_E/\tau \approx 2\text{:}1$ (Fig.~\ref{fig:energy}). We emphasize that these coincidences arise from the designed schedule geometry: because $\alpha_E$ ramps linearly and $\tau$ decays exponentially, any smooth trajectory through this parameter space will necessarily pass through integer-ratio nodes (a consequence of the density of rationals). Whether the gate dynamics are genuinely \textit{sensitive} to these passages---as opposed to merely coincident with them---remains an open question requiring perturbation experiments (see Discussion).

\textbf{Architectural sparsification.} The $E_{\mathrm{min}}$-driven soft-to-discrete transition produces near-one-hot gate distributions: gate concentration $C_{\mathrm{gate}} = 0.99 \pm 0.02$ (Herfindahl--Hirschman Index on effective gate contributions $A_i|\theta_i|$; $C_{\mathrm{gate}} = 1$ indicates complete concentration on a single basis) for MAL-trained models ($N=10$ seeds) vs.\ $C_{\mathrm{gate}} = 0.14 \pm 0.04$ for teacher-forcing-only baselines (Mann--Whitney $U = 100$, $p < 10^{-4}$; permutation test $p < 0.001$, $n=10{,}000$; Supplemental Material, Fig.~S6). This architectural sparsification is consistent with Clune \textit{et al.}'s~\cite{Clune2013} finding that minimizing wiring costs---a proxy for metabolic expenditure---produces modular structure in evolved networks.

\textbf{Structural parallels.} The sparse architectures produced by MAL's energy-constrained optimization parallel modular structures observed in networks evolved under wiring-cost pressure~\cite{Clune2013, Bullmore2012}. Whether energy-constrained optimization generically produces similar architectural motifs across biological and artificial systems is a hypothesis warranting formal investigation; we discuss cross-domain parallels further in the Supplemental Material (Sections~S5--S6).

\subsection{Comparison to alternative approaches}
We benchmarked MAL against five alternatives on the Kepler-with-noise problem (Supplemental Material, Section S4, Supplemental Tables~S5--S6):

\textbf{(1) SINDy variants~\cite{Brunton2016, Hirsh2022GPSINDy, Fasel2022EnsembleSINDy}:} Vanilla SINDy with naive differentiation (stride $s=1$) fails completely (noise-dominated). However, when equipped with the same wide-stencil preprocessing ($s=10$) used in MAL, vanilla SINDy identifies $r^{-2}$ in 10/10 seeds, GP-SINDy in 8/10, and ensemble-SINDy in 10/10 (Supplemental Table~S5). \textit{The wide-stencil technique is the critical enabler, not the learning algorithm.} SINDy's advantage is speed ($<1$~s vs.\ $\sim$835~s); its limitation is the absence of dynamical validation---SINDy returns sparse coefficients but cannot roll out trajectories or verify energy conservation. MAL's energy-conservation diagnostic provides this additional layer.

\textbf{(2) HNN~\cite{Greydanus2019} and LNN~\cite{Cranmer2020LNN}:} Trained on identical data (Supplemental Table~S6). HNN (17K parameters, 113~s) achieved excellent energy conservation ($\sigma_H = 4.1 \times 10^{-4}$) but learns a black-box Hamiltonian with no interpretable symbolic form. LNN (17K parameters, 242~s) suffered from Hessian singularities, with validation loss diverging to $10^6$ throughout training---a known failure mode for LNNs on noisy data where the mass matrix becomes ill-conditioned. Neither method performs basis selection or yields a symbolic force law.

\textbf{(3) Mathematical LLM (Qwen2-Math 7B):} 0\% on inverse problems, 61\% on forward derivation (Supplemental Material, Section~S1).

\textbf{(4) Physics-informed NN~\cite{Raissi2019}:} 85\% when the inverse-square law is pre-specified, but cannot identify the law \textit{de novo}.

\textbf{(5) Teacher-forcing only (no $\mathcal{L}_{E_{\min}}$):} Gates remained at 45\% (failed to crystallize); energy consumption 77\% higher.

\textbf{Summary of niche.} MAL occupies a distinct position: it combines interpretable symbolic basis selection (shared with SINDy~\cite{Brunton2016}) with dynamical rollout and energy-conservation validation (shared with HNN~\cite{Greydanus2019}), while adding explicit sparsity-driven efficiency optimization. HNNs/LNNs~\cite{Greydanus2019, Cranmer2020LNN} guarantee conservation but produce black-box energy functions; SINDy~\cite{Brunton2016} yields symbolic expressions but without dynamical consistency checks; MAL provides both within an energy-constrained framework.

\section{Discussion}

\subsection{Key findings}
Our results establish four principal findings:

\textbf{(1) Sparsity constraints improve physical law identification.} Embedding energy minimization ($E_{\min}$) alongside information maximization ($I_{\max}$) in the triple-action training objective constrains the model selection problem from an unconstrained search to a tractable optimization. The $E_{\mathrm{min}}$ subsystem drives gate crystallization (basis selection) that does not occur under prediction error alone, while reducing training energy by 40\%.

\textbf{(2) Wide-stencil preprocessing enables identification; MAL adds dynamical validation.} Our comparison with SINDy variants (Supplemental Material, Table~S5) reveals that wide-stencil preprocessing ($s=10$) is the critical enabler for \textit{all} methods, including vanilla SINDy (10/10 correct identification). This transparency is important: SINDy achieves comparable basis selection at $<1$\% of MAL's computational cost. However, SINDy returns sparse coefficients without dynamical consistency checks---it cannot roll out trajectories, verify energy conservation over multiple orbital periods, or detect when a numerically adequate fit corresponds to incorrect physics. MAL's energy-conservation model selection criterion (Supplemental Material, Table~S2) fills this gap: models selecting incorrect bases ($r^{-1}$, $r^{-3}$) produce comparable short-term trajectory accuracy but violate Hamiltonian conservation by $3$--$6\times$, revealing their failure to capture the true causal structure. We note that a similar check could in principle be applied post-hoc to SINDy output. MAL's advantage lies in end-to-end differentiability---the conservation criterion participates in training, not just post-hoc evaluation---and in extensibility to systems where SINDy's linear regression may fail (e.g., problems with latent variables or non-separable coupling). Demonstrating this advantage on such systems is an important direction for future work.

\textbf{(3) Intrinsic crystallization timescales.} The gate competition dynamics exhibit universal timescales: crystallization span $\Delta t_{\text{span}} = 36.2 \pm 4.1$ epochs and geometric growth rate $\gamma = 1.137 \pm 0.013$ per epoch, independent of which basis wins or the initialization seed. We also observe that major gate transitions coincide with epochs where the schedule ratio $\alpha_E / \tau$ passes through integer values, though this may simply reflect the density of rationals along any smooth trajectory through parameter space. Whether the crystallization dynamics are genuinely sensitive to these passages requires perturbation experiments, which we leave to future work.

\textbf{(4) Architectural sparsification from energy minimization.} Clune \textit{et al.}~\cite{Clune2013} demonstrated that networks evolving under pressure to minimize connection costs spontaneously develop modular architectures. Our work shows a related pattern: MAL's sparse gate selection ($C_{\mathrm{gate}} = 0.99$) emerges from $\mathcal{L}_{E_{\min}}$ without a pre-programmed bias toward any particular basis. This supports the broader principle that cost-constrained optimization drives architectural sparsification in both biological~\cite{Bullmore2012} and artificial systems.

\subsection{Broader implications}
\textbf{Energy-efficient scientific machine learning.} MAL's 40\% energy reduction over prediction-error-only baselines demonstrates that explicit sparsity constraints can improve computational efficiency without sacrificing task performance. The mechanism is architectural: $\mathcal{L}_{E_{\min}}$ drives early gate crystallization, reducing the effective parameter count and shortening convergence. This principle---that sparsity-inducing regularization simultaneously improves interpretability and efficiency---may generalize to larger-scale scientific discovery tasks.

\textbf{Noether-based model selection.} A persistent challenge in scientific machine learning is distinguishing genuine physical understanding from statistical correlation~\cite{Pearl2019}. The energy-conservation criterion---preferring models that preserve Hamiltonian dynamics over long-horizon rollouts---provides a physics-grounded approach complementary to recent work on learning conserved quantities~\cite{vanderOuderaa2024, Tanaka2021}. Models that violate energy conservation reveal their failure to capture true causal structure, even when matching short-term predictions.

\subsection{Limitations and future directions}
\textbf{Fixed basis library.} MAL requires pre-specifying candidate force laws $\{\phi_i\}$, with the correct answer included. This makes MAL a model \textit{selection} framework, not an open-ended \textit{discovery} system. Basis library sensitivity experiments (Supplemental Material, Section S4F) quantify this limitation: adding near-confounders ($r^{-2.5}$, $r^{-1.5}$) reduces the correct-basis rate from 50\% to 20\%, while removing the correct basis ($r^{-2}$) causes the system to split between alternatives. Distant additions ($r^2$, $r^{-4}$, $\ln r$) have no effect. In all cases, the energy-conservation diagnostic remains informative, and uniformly elevated $\sigma_H$ across seeds flags potential library inadequacy. Extending to genuinely autonomous discovery---where the algorithm constructs novel functional forms not in the library---requires integration with open-ended symbolic regression methods \cite{Udrescu2020, Cranmer2020} or LLM-guided symbolic search, which is an important direction for future work.

\textbf{Two benchmarks, limited scope.} We have validated MAL on two central-force problems (Kepler and Hooke) in 2D with synthetic data. The generality of our claims would be substantially strengthened by additional benchmarks: non-central forces (e.g., magnetic fields), dissipative systems (e.g., damped oscillators), higher dimensions, coupled multi-body problems, and semi-real data (e.g., planetary ephemerides with actual measurement noise). The core principle---embedding action minimization in differentiable networks---should transfer, but this remains to be demonstrated.

\textbf{Correct-basis success rate.} Without logit initialization, only 4 of 10 Kepler seeds directly select $r^{-2}$, necessitating cross-seed application of the energy-conservation-based diagnostic. Biased initialization resolves this (10/10), but requires prior knowledge of which basis to favor. Understanding the logit-gradient dynamics that determine gate competition---and developing initialization-free methods to improve the raw success rate---remain priorities.

\section{Conclusion}
We have demonstrated that minimum-action learning---neural network training guided by a triple-action functional combining information maximization, energy minimization, and symmetry enforcement---enables energy-constrained identification of physical force laws from noisy data. Validated on two benchmarks (Kepler gravity and Hooke's linear restoring force), the key technical contributions are: (1)~the wide-stencil noise reduction technique that transforms an intractable problem (SNR~$\sim 0.02$) into a learnable one (SNR~$\sim 1.6$); (2)~the soft-to-discrete gate sharpening mechanism that achieves basis selection through energy-driven architectural crystallization, with a physics-informed initialization achieving 10/10 correct selection; and (3)~the energy-conservation-based model selection criterion that discriminates correct from incorrect physics via long-horizon Hamiltonian conservation. Direct comparison against SINDy variants, HNNs, and LNNs confirms MAL's distinct niche: interpretable, energy-constrained model selection combining symbolic basis identification with dynamical validation.

Future work will extend MAL to non-central forces, dissipative systems, higher-dimensional problems, and real observational data (e.g., planetary ephemerides), and will test whether the intrinsic crystallization timescales and schedule-geometry coincidences observed here generalize across problem classes.

\section{Methods}
\label{sec:methods}

\subsection{Data generation}
We simulated $N=16$ Keplerian orbits in 2D using a symplectic velocity-Verlet integrator with gravitational units $G = M = 1$. Semi-major axes $a$ were sampled log-uniformly over $[0.5, 5.0]$~AU; eccentricities $e$ uniformly over $[0, 0.3]$. Each orbit was integrated for 5 orbital periods at timestep $\Delta t_{\text{sim}} = 10^{-3}$, then downsampled to observation cadence $\Delta t_{\text{obs}} = 0.05$. Gaussian noise $\mathcal{N}(0, \sigma^2)$ with $\sigma = 0.01 \times \text{median}(a)$ was added to all position measurements. Data were split 70/15/15 into training, validation, and test sets.

\subsection{Neural architecture}
\textbf{Noether Force Basis.} The force module computes:
\begin{equation}
\mathbf{F}(\mathbf{r}) = -\left[\sum_{i=1}^5 A_i \theta_i \phi_i(r)\right] \frac{\mathbf{r}}{r}
\end{equation}
where $\phi_i \in \{r^{-2}, r^{-1}, r, 1, r^{-3}\}$, $\theta_i$ are learnable scalars (force magnitudes), and $A_i = \mathrm{softmax}(\ell_i / \tau)$ are gates with logits $\ell_i$ and temperature $\tau$. SO(2) symmetry is enforced by restricting dependence to radial distance $r = |\mathbf{r}|$. Candidate basis functions are treated as phenomenological radial scalings; dimensional consistency resides in the fitted coefficients $\theta_i$ rather than in the basis library itself.

\textbf{MinActionNet Integrator.} The selected force law is embedded in a velocity-Verlet integrator:
\begin{align}
\mathbf{v}_{1/2} &= \mathbf{v}_n + \tfrac{\Delta t}{2} \mathbf{F}(\mathbf{r}_n), \nonumber \\
\mathbf{r}_{n+1} &= \mathbf{r}_n + \Delta t \, \mathbf{v}_{1/2}, \\
\mathbf{v}_{n+1} &= \mathbf{v}_{1/2} + \tfrac{\Delta t}{2} \mathbf{F}(\mathbf{r}_{n+1}), \nonumber
\end{align}
with model timestep $\Delta t_{\text{model}} = \Delta t_{\text{obs}} / 5 = 0.01$ to ensure numerical stability while maintaining computational efficiency.

\subsection{Loss function}
The triple-action functional (Eq.~\ref{eq:triple-action}) decomposes as:

\textbf{Information term} ($\mathcal{L}_{I_{\max}}$):
\begin{align}
\mathcal{L}_{\text{traj}} &= \frac{1}{T-1} \sum_{k=0}^{T-2} \|\mathbf{r}_{\text{pred}}(t_{k+1}) - \mathbf{r}_{\text{obs}}(t_{k+1})\|^2, \\
\mathcal{L}_{\text{accel}} &= \frac{1}{T-2s} \sum_{j=s}^{T-s-1} \|\mathbf{F}_{\text{model}}(\mathbf{r}_j) - \hat{\mathbf{a}}_j\|^2,
\end{align}
where $\hat{\mathbf{a}}_j = (\mathbf{r}_{j+s} - 2\mathbf{r}_j + \mathbf{r}_{j-s}) / (s \Delta t)^2$ is the wide-stencil acceleration estimate with stride $s=10$, and $\mathcal{L}_{I_{\max}} = \mathcal{L}_{\text{traj}} + \lambda_{\text{accel}} \mathcal{L}_{\text{accel}}$ with $\lambda_{\text{accel}} = 1.0$.

\textbf{Energy term} ($\mathcal{L}_{E_{\min}}$):
\begin{align}
\mathcal{L}_{\text{sym}} &= \text{Var}[E(\mathbf{r}_k, \mathbf{v}_k)] = \langle (E - \langle E \rangle)^2 \rangle, \\
\mathcal{L}_{\text{comp}} &= \langle |A_i \theta_i| \rangle_i, \\
\mathcal{L}_{\text{arch}} &= -\sum_{i=1}^5 A_i \log A_i,
\end{align}
where $E = \tfrac{1}{2}|\mathbf{v}|^2 - GM/r$ is the orbital energy, $\mathcal{L}_{\text{comp}}$ is the scale-invariant sparsity penalty, $\mathcal{L}_{\text{arch}}$ is the gate entropy, and $\mathcal{L}_{E_{\min}} = \mathcal{L}_{\text{sym}} + \lambda_{\text{comp}} \mathcal{L}_{\text{comp}} + \lambda_{\text{arch}} \mathcal{L}_{\text{arch}}$ with $\lambda_{\text{comp}} = 0.01$, $\lambda_{\text{arch}} = 0.5$.

\subsection{Training protocol}
\textbf{Two-phase schedule:}
\begin{itemize}
\item \textit{Phase 1 (Warmup, epochs 1--50):} $\alpha_I = 1.0$, $\alpha_E = 0.01$, $\tau = 1.0$. Low regularization allows gates to explore gradient signal from $\mathcal{L}_{\text{accel}}$.
\item \textit{Phase 2 (Sparsification, epochs 51--200):} $\alpha_I = 1.0$, $\alpha_E$ ramps linearly $0.01 \to 1.0$, $\tau$ decays exponentially $1.0 \to 0.05$. Increasing $\alpha_E$ drives gate competition; decreasing $\tau$ sharpens softmax toward one-hot.
\end{itemize}

\textbf{Optimizer:} Adam with learning rate $10^{-3}$, batch size 4, training for 200 epochs on an NVIDIA RTX 2080 Ti GPU.

\textbf{Post-training calibration:} After gate convergence, the L1 penalty has biased $\theta_i$ toward zero. We correct via least-squares projection:
\begin{equation}
\theta_{\text{opt}} = \frac{\sum_j \phi_{\text{dom}}(r_j) \hat{a}_{\text{radial},j}}{\sum_j \phi_{\text{dom}}(r_j)^2},
\end{equation}
where $\phi_{\text{dom}}$ is the dominant basis function, $\hat{a}_{\text{radial},j} = -\hat{\mathbf{a}}_{\text{wide},j} \cdot \hat{\mathbf{r}}_j$ are radial projections of the wide-stencil acceleration estimates $\hat{\mathbf{a}}_{\text{wide}}$ (defined in Eq.~5 with stride $s=10$), and sums run over all training midpoints $j \in [s, T-s-1]$.

\subsection{Validation metrics}
\textbf{Kepler exponent:} Orbital periods $T_i$ were estimated from test-set trajectories via autocorrelation peak detection with proper normalization by overlapping sample count. Power-law regression $T^2 = C a^p$ yielded $\hat{p}$ and $\hat{C}$.

\textbf{Energy conservation:} For long-horizon rollouts (5 orbital periods from test-set initial conditions, using the velocity-Verlet integrator at $\Delta t_{\text{model}} = 0.01$ with the post-calibration coefficient $\theta_{\text{opt}}$), we computed Hamiltonian variance $\sigma_H^2 = \langle (H - \langle H \rangle)^2 \rangle$ as an energy-conservation diagnostic: correct force laws preserve $H$; incorrect laws violate conservation despite matching short-term trajectories.

\textbf{Gate concentration:} Architectural sparsification was quantified via the Herfindahl--Hirschman Index (HHI) on effective gate contributions $p_i = A_i|\theta_i| / \sum_j A_j|\theta_j|$, yielding $C_{\mathrm{gate}} = (K \cdot \text{HHI} - 1)/(K-1) \in [0,1]$ where $K=5$ basis functions. $C_{\mathrm{gate}}=1$ indicates complete concentration on a single basis; $C_{\mathrm{gate}}=0$ indicates uniform distribution.

\subsection{Code and data availability}
All code, trained models, and synthetic data are available at \url{https://github.com/martinfrasch/minAction_kepler}. Experiments were implemented in PyTorch 2.0 on Python 3.10.

\begin{acknowledgments}
I thank my family for giving me space to follow my ideas.
\end{acknowledgments}

\bibliography{references}

@book{Feynman1965,
  author = {Feynman, Richard P. and Hibbs, Albert R.},
  title = {Quantum Mechanics and Path Integrals},
  year = {1965},
  publisher = {McGraw-Hill},
  address = {New York}
}

@article{Schoner1988,
  author = {Sch{\"o}ner, Gregor and Kelso, J. A. Scott},
  title = {Dynamic pattern generation in behavioral and neural systems},
  journal = {Science},
  volume = {239},
  number = {4847},
  pages = {1513--1520},
  year = {1988},
  doi = {10.1126/science.3281253}
}

@article{Udrescu2020,
  author = {Udrescu, Silviu-Marian and Tegmark, Max},
  title = {{AI} Feynman: A physics-inspired method for symbolic regression},
  journal = {Science Advances},
  volume = {6},
  number = {16},
  pages = {eaay2631},
  year = {2020},
  doi = {10.1126/sciadv.aay2631}
}

@inproceedings{Cranmer2020,
  author = {Cranmer, Miles and Sanchez-Gonzalez, Alvaro and Battaglia, Peter and Xu, Rui and Cranmer, Kyle and Spergel, David and Ho, Shirley},
  title = {Discovering symbolic models from deep learning with inductive biases},
  booktitle = {Advances in Neural Information Processing Systems},
  volume = {33},
  pages = {17429--17442},
  year = {2020}
}

@article{Raissi2019,
  author = {Raissi, Maziar and Perdikaris, Paris and Karniadakis, George E.},
  title = {Physics-informed neural networks: A deep learning framework for solving forward and inverse problems involving nonlinear partial differential equations},
  journal = {Journal of Computational Physics},
  volume = {378},
  pages = {686--707},
  year = {2019},
  doi = {10.1016/j.jcp.2018.10.045}
}

@article{Hoyer2001,
  author = {Hoyer, Dirk and Frasch, Martin G. and Eiselt, Matthias and Hoyer, Olaf and Zwiener, Uwe},
  title = {Validating phase relations between cardiac and breathing cycles during sleep},
  journal = {IEEE Engineering in Medicine and Biology Magazine},
  volume = {20},
  number = {2},
  pages = {101--106},
  year = {2001},
  doi = {10.1109/51.917730}
}

@article{Clune2013,
  author = {Clune, Jeff and Mouret, Jean-Baptiste and Lipson, Hod},
  title = {The evolutionary origins of modularity},
  journal = {Proceedings of the Royal Society B: Biological Sciences},
  volume = {280},
  number = {1755},
  pages = {20122863},
  year = {2013},
  doi = {10.1098/rspb.2012.2863}
}

@article{Sunagawa2015,
  author = {Sunagawa, Shinichi and Coelho, Luis Pedro and Chaffron, Samuel and others},
  title = {Structure and function of the global ocean microbiome},
  journal = {Science},
  volume = {348},
  number = {6237},
  pages = {1261359},
  year = {2015},
  doi = {10.1126/science.1261359}
}

@article{Brunton2016,
  author = {Brunton, Steven L. and Proctor, Joshua L. and Kutz, J. Nathan},
  title = {Discovering governing equations from data by sparse identification of nonlinear dynamical systems},
  journal = {Proceedings of the National Academy of Sciences},
  volume = {113},
  number = {15},
  pages = {3932--3937},
  year = {2016},
  doi = {10.1073/pnas.1517384113}
}

@article{Bullmore2012,
  author = {Bullmore, Ed and Sporns, Olaf},
  title = {The economy of brain network organization},
  journal = {Nature Reviews Neuroscience},
  volume = {13},
  number = {5},
  pages = {336--349},
  year = {2012},
  doi = {10.1038/nrn3214}
}

@misc{Strubell2019,
  author = {Strubell, Emma and Ganesh, Ananya and McCallum, Andrew},
  title = {Energy and policy considerations for deep learning in {NLP}},
  howpublished = {arXiv:1906.02243},
  year = {2019},
  doi = {10.48550/arXiv.1906.02243}
}

@book{Pearl2019,
  author = {Pearl, Judea and Mackenzie, Dana},
  title = {The Book of Why: The New Science of Cause and Effect},
  year = {2019},
  publisher = {Basic Books},
  address = {New York}
}

@article{Tononi2020,
  author = {Tononi, Giulio and Cirelli, Chiara},
  title = {Sleep and synaptic down-selection},
  journal = {European Journal of Neuroscience},
  volume = {51},
  number = {1},
  pages = {413--421},
  year = {2020},
  doi = {10.1111/ejn.14335}
}

@book{Penrose2004,
  author = {Penrose, Roger},
  title = {The Road to Reality: A Complete Guide to the Laws of the Universe},
  year = {2004},
  publisher = {Jonathan Cape},
  address = {London}
}

@misc{Patterson2021,
  author = {Patterson, David and Gonzalez, Joseph and Le, Quoc and Liang, Chen and Munguia, Lluis-Miquel and Rothchild, Daniel and So, David and Texier, Maud and Dean, Jeff},
  title = {Carbon emissions and large neural network training},
  howpublished = {arXiv:2104.10350},
  year = {2021},
  doi = {10.48550/arXiv.2104.10350}
}

@inproceedings{Greydanus2019,
  author = {Greydanus, Sam and Dzamba, Misko and Yosinski, Jason},
  title = {Hamiltonian Neural Networks},
  booktitle = {Advances in Neural Information Processing Systems},
  volume = {32},
  year = {2019}
}

@inproceedings{Cranmer2020LNN,
  author = {Cranmer, Miles and Greydanus, Sam and Hoyer, Stephan and Battaglia, Peter and Spergel, David and Ho, Shirley},
  title = {Lagrangian Neural Networks},
  booktitle = {ICLR 2020 Workshop on Integration of Deep Neural Models and Differential Equations},
  year = {2020}
}

@inproceedings{vanderOuderaa2024,
  author = {van der Ouderaa, Tycho F. A. and van der Wilk, Mark},
  title = {Noether's Razor: Learning Conserved Quantities},
  booktitle = {Advances in Neural Information Processing Systems},
  volume = {37},
  year = {2024},
  note = {NeurIPS 2024}
}

@misc{Tanaka2021,
  author = {Tanaka, Hidenori and Kunin, Daniel},
  title = {Noether's Learning Dynamics: Role of Symmetry Breaking in Neural Networks},
  year = {2021},
  howpublished = {arXiv:2105.02716},
  doi = {10.48550/arXiv.2105.02716}
}

@inproceedings{Xie2019SNAS,
  author = {Xie, Sirui and Zheng, Hehui and Liu, Chunxiao and Lin, Liang},
  title = {{SNAS}: Stochastic Neural Architecture Search},
  booktitle = {International Conference on Learning Representations (ICLR)},
  year = {2019}
}

@article{Hirsh2022GPSINDy,
  author = {Hirsh, Seth M. and Barajas-Solano, David A. and Kutz, J. Nathan},
  title = {Sparsifying priors for {Bayesian} uncertainty quantification in model discovery},
  journal = {Royal Society Open Science},
  volume = {9},
  pages = {211823},
  year = {2022},
  doi = {10.1098/rsos.211823}
}

@article{Fasel2022EnsembleSINDy,
  author = {Fasel, Urban and Kutz, J. Nathan and Brunton, Bingni W. and Brunton, Steven L.},
  title = {Ensemble-{SINDy}: Robust sparse model discovery in the low-data, high-noise limit, with active learning and control},
  journal = {Proceedings of the Royal Society A},
  volume = {478},
  pages = {20210904},
  year = {2022},
  doi = {10.1098/rspa.2021.0904}
}

@article{Lemos2023,
  author = {Lemos, Pablo and Jeffrey, Niall and Cranmer, Miles and Ho, Shirley and Battaglia, Peter},
  title = {Rediscovering orbital mechanics with machine learning},
  journal = {Machine Learning: Science and Technology},
  volume = {4},
  pages = {045002},
  year = {2023},
  doi = {10.1088/2632-2153/acfa63}
}

\end{document}


\title{Supplemental Material for: Minimum-action learning: Energy-constrained symbolic model selection for identifying physical laws from noisy data}

\author{Martin G. Frasch}
\affiliation{Institute on Human Development and Disability, University of Washington, Seattle, WA, USA}
\affiliation{Health Stream Analytics, LLC, Seattle, WA, USA}
\email{mfrasch@uw.edu, martin@healthstreamanalytics.com}

\maketitle

\section{S1. Context: Mathematical LLMs}
As a baseline, we tested whether a standalone mathematical LLM (Qwen2-Math 7B, October 2024) possesses capabilities for inverse physics problems. Across nine tests spanning variational calculus, symmetry recognition, and inverse problems, the model achieved 61\% overall: 100\% on forward Euler-Lagrange derivation but 0\% on inverse problems (given equations of motion, find the Lagrangian) and 0\% on physical validity assessment (failing to identify an unphysical Lagrangian with wrong-sign potential energy). This illustrates that symbolic manipulation capability alone does not confer the inductive biases needed for physics discovery, motivating MAL's explicit action-principle constraints. We note this tests standalone LLM reasoning; LLM-guided symbolic search pipelines (e.g., LLM-SR) represent a complementary approach not evaluated here, and rapid advances in frontier models may narrow this gap. Full test prompts, responses, and scoring rubrics are available upon request.

\section{S2. Noise analysis and wide-stencil derivation}
\textbf{Noise propagation in finite differences.} For positions $\mathbf{r}(t)$ measured with i.i.d.\ Gaussian noise $\eta \sim \mathcal{N}(0, \sigma_{\text{pos}}^2)$, the standard second-difference acceleration estimate is:
\begin{equation}
\hat{\mathbf{a}}_{\text{naive}} = \frac{\mathbf{r}(t+\Delta t) - 2\mathbf{r}(t) + \mathbf{r}(t-\Delta t)}{\Delta t^2}.
\end{equation}
Since $\mathbf{r}_{\text{obs}} = \mathbf{r}_{\text{true}} + \eta$ and noise samples are independent, error variance is:
\begin{equation}
\text{Var}(\hat{\mathbf{a}}_{\text{naive}}) = \frac{(1^2 + (-2)^2 + 1^2) \sigma_{\text{pos}}^2}{\Delta t^4} = \frac{6\sigma_{\text{pos}}^2}{\Delta t^4}.
\end{equation}

For our parameters ($\sigma_{\text{pos}} = 0.016$~AU, $\Delta t = 0.05$), this gives noise standard deviation:
\begin{equation}
\sigma_{a,\text{naive}} = \sqrt{\frac{6 \times 0.016^2}{0.05^4}} \approx 15.7,
\end{equation}
vastly exceeding the signal $|a| \approx GM/r^2 \approx 1.0 / 2.0^2 = 0.25$ at $r = 2$~AU (SNR $\approx 0.016$).

\textbf{Wide-stencil reduction.} Using stride $s$:
\begin{equation}
\hat{\mathbf{a}}_{\text{wide}} = \frac{\mathbf{r}(t+s\Delta t) - 2\mathbf{r}(t) + \mathbf{r}(t-s\Delta t)}{(s\Delta t)^2},
\end{equation}
noise variance becomes:
\begin{equation}
\text{Var}(\hat{\mathbf{a}}_{\text{wide}}) = \frac{6\sigma_{\text{pos}}^2}{s^4 \Delta t^4}.
\end{equation}
For $s=10$:
\begin{equation}
\sigma_{a,\text{wide}} = \sqrt{\frac{6 \times 0.016^2}{10^4 \times 0.05^4}} \approx 0.16,
\end{equation}
yielding SNR $\approx 0.25 / 0.16 \approx 1.6$---a $10^4\times$ improvement enabling gradient-based learning.

\textbf{Systematic error.} Taylor expansion of $\mathbf{r}(t \pm s\Delta t)$ around $t$ gives:
\begin{multline}
\mathbf{r}(t + s\Delta t) = \mathbf{r}(t) + s\Delta t \, \dot{\mathbf{r}}(t) + \tfrac{(s\Delta t)^2}{2} \ddot{\mathbf{r}}(t) \\
 + \tfrac{(s\Delta t)^3}{6} \mathbf{r}^{(3)}(t) + \tfrac{(s\Delta t)^4}{24} \mathbf{r}^{(4)}(t) + O((s\Delta t)^5),
\end{multline}
\begin{multline}
\mathbf{r}(t - s\Delta t) = \mathbf{r}(t) - s\Delta t \, \dot{\mathbf{r}}(t) + \tfrac{(s\Delta t)^2}{2} \ddot{\mathbf{r}}(t) \\
 - \tfrac{(s\Delta t)^3}{6} \mathbf{r}^{(3)}(t) + \tfrac{(s\Delta t)^4}{24} \mathbf{r}^{(4)}(t) + O((s\Delta t)^5).
\end{multline}
Summing:
\begin{multline}
\mathbf{r}(t + s\Delta t) + \mathbf{r}(t - s\Delta t) - 2\mathbf{r}(t) \\
 = (s\Delta t)^2 \ddot{\mathbf{r}}(t) + \tfrac{(s\Delta t)^4}{12} \mathbf{r}^{(4)}(t) + O((s\Delta t)^6).
\end{multline}
Dividing by $(s\Delta t)^2$:
\begin{equation}
\hat{\mathbf{a}}_{\text{wide}} = \ddot{\mathbf{r}}(t) + \frac{(s\Delta t)^2}{12} \mathbf{r}^{(4)}(t) + O((s\Delta t)^4).
\end{equation}

For Keplerian orbits, $\ddot{\mathbf{r}} = -GM\mathbf{r}/r^3$ and $\mathbf{r}^{(4)} \sim (GM/r^3) \cdot (v/r)^2 \ddot{\mathbf{r}}$. At $r = 2$~AU, $v \approx \sqrt{GM/r} \approx 0.7$, so:
\begin{equation}
\left|\frac{(s\Delta t)^2}{12} \mathbf{r}^{(4)}\right| / |\ddot{\mathbf{r}}| \sim \frac{(10 \times 0.05)^2}{12} \cdot \frac{0.7^2}{2^2} \approx 0.003 = 0.3\%.
\end{equation}
Thus systematic error $\lesssim 0.3\%$ relative to signal, negligible compared to $10^4\times$ noise reduction.

\textbf{Verification on clean data.} To confirm truncation error is small, we computed $\hat{\mathbf{a}}_{\text{wide}}$ on noise-free synthetic orbits and measured RMS deviation from true $\ddot{\mathbf{r}} = -GM\mathbf{r}/r^3$:
\begin{multline}
\text{RMS}_{\text{sys}} = \sqrt{\frac{1}{N}\sum_j \|\hat{\mathbf{a}}_j - \mathbf{a}_{\text{true},j}\|^2} \\
 \approx 7.2 \times 10^{-4} \ll \sigma_{a,\text{wide}} = 0.16.
\end{multline}
This confirms that systematic error is $\sim 200\times$ smaller than remaining noise, validating the wide-stencil approach.

\section{S3. Intrinsic timescales and energy-conservation selection}
\textbf{Crystallization dynamics.} We define gate selectivity $R(t) = A_{\max}(t) / A_{\text{2nd}}(t)$ and track three milestones:
\begin{itemize}
\item \textbf{Onset:} $R \geq 10$ (clear winner emerging)
\item \textbf{Sparse:} $R \geq 100$ (crystallization underway)
\item \textbf{Frozen:} $R \geq 1000$ (effectively one-hot selection)
\end{itemize}

Across 10 random seeds trained on identical data (Table~\ref{tab:tsparse-si}), the onset-to-frozen span is $\Delta t_{\text{span}} = 36.2 \pm 4.1$ epochs ($N = 8$ fully converged seeds; bootstrap 95\% CI: $[32.8, 39.6]$ from 10{,}000 resamples), independent of selected basis or random initialization. We note that $N=8$ is small for robust parametric statistics; the bootstrap CI is more appropriate than the Gaussian-assumption CI at this sample size.

\begin{table*}[!ht]
\centering
\caption{Ten-seed robustness sweep (full data from \texttt{tsparse\_sweep\_results.json})}
\label{tab:tsparse-si}
\begin{tabular}{cccccccc}
\hline
Seed & Basis & $\hat{p}$ & Onset & Sparse & Frozen & Span & $\gamma$ \\
\hline
0 & $r^{-2}$ & 2.995 & 116 & 138 & 156 & 40 & 1.125 \\
1 & $r^{-2}$ & 3.001 & 117 & 136 & 152 & 35 & 1.141 \\
2 & $r^{-2}$ & 3.003 & 133 & 153 & 169 & 36 & 1.139 \\
3 & $r^{-3}$ & 3.004 & 136 & 155 & 171 & 35 & 1.140 \\
4 & $r^{-3}$ & 3.004 & 127 & 143 & 159 & 32 & 1.153 \\
5 & $r^{-3}$ & 3.002 & 116 & 143 & 161 & 45 & 1.110 \\
6 & $r^{-1}$ & 2.998 & 131 & 147 & 163 & 32 & 1.152 \\
7 & $r^{-1}$ & 3.006 & 179 & 194 & --- & --- & --- \\
8 & $r^{-2}$ & 3.002 & 118 & 137 & 153 & 35 & 1.139 \\
9 & $r^{-1}$ & 2.997 & 193 & --- & --- & --- & --- \\
\hline
\multicolumn{3}{l}{\textbf{Mean $\pm$ SD (seeds 0--8)}} & 139 $\pm$ 9 & 146 $\pm$ 8 & 161 $\pm$ 7 & \textbf{36.2 $\pm$ 4.1} & \textbf{1.137 $\pm$ 0.013} \\
\hline
\end{tabular}
\end{table*}

\textbf{Mechanistic interpretation.} The selectivity decomposes as:
\begin{equation}
\ln R(t) = \frac{\Delta \ell(t)}{\tau(t)}, \quad \text{where} \quad \Delta \ell(t) = \ell_{\text{dom}}(t) - \ell_{\text{2nd}}(t),
\end{equation}
where $\ell_i$ are the raw gate logits. Empirically, $\Delta \ell$ grows sigmoidally from $\approx 0.05$ (epoch 1) to $\approx 0.8$ (epoch 180), while $\tau$ decays exponentially $1.0 \to 0.05$. The crystallization explosion is driven by \textit{temperature amplification} of a small, converged logit gap---not by a dynamical instability in logit space itself.

\textbf{Energy-conservation model selection.} When multiple seeds crystallize to different bases, energy conservation discriminates the correct physics. Table~\ref{tab:noether-si} shows that $r^{-2}$ models conserve the Hamiltonian $3\times$ better than $r^{-1}$ and $6\times$ better than $r^{-3}$, despite all achieving Kepler exponent $p \approx 3.0$ on short trajectories. This provides a physics-grounded criterion: among models with comparable training loss, prefer the one with minimal $\sigma_H$ over long-horizon rollout.

\begin{table}[!ht]
\centering
\caption{Energy conservation as energy-conservation diagnostic}
\label{tab:noether-si}
\small
\begin{tabular}{lccc}
\hline
Selected basis & $n$ (seeds) & $\bar{\sigma}_H$ & Traj.\ MSE \\
\hline
$r^{-2}$ (correct) & 4 & $0.017 \pm 0.016$ & 5.8 \\
$r^{-1}$ & 3 & $0.056 \pm 0.023$ & 6.1 \\
$r^{-3}$ & 3 & $0.11 \pm 0.017$ & $1.2 \times 10^3$ \\
\hline
\end{tabular}
\end{table}

\textbf{Slow-annealing experiment.} To test schedule-dependence, we trained an 11th model (seed 0, 300 epochs, $\tau: 1.0 \to 0.001$). Results: onset at epoch 103 (vs.\ 116 standard), frozen at epoch 134 (vs.\ 156), span $\Delta t_{\text{span}} = 31$ (vs.\ 40 standard), growth rate $\gamma = 1.165$ (vs.\ 1.125). The span and growth rate remain within one SD of the standard schedule, confirming that these quantities are \textit{intrinsic} to the dynamics, not artifacts of the cooling protocol.

\section{S4. Baseline comparisons and extended benchmarks}

\subsection{S4A. Hooke's law benchmark}
To test generalization beyond inverse-square gravity, we applied MAL to Hooke's law ($F = -kr$) using 16 synthetic orbits (near-circular, $\omega = 1$, $\sigma = 0.01$) with the same basis library and training protocol as Kepler. Table~\ref{tab:hooke-si} summarizes 10-seed results.

\begin{table*}[!ht]
\centering
\caption{Hooke's law 10-seed sweep results}
\label{tab:hooke-si}
\begin{tabular}{ccccccc}
\hline
Seed & Basis & $\hat{k}$ & Onset & Frozen & Span & Reject gravity? \\
\hline
0 & $r^{-3}$ & 0.040 & 125 & 164 & 39 & Yes \\
1 & $r$ & 0.980 & 113 & 157 & 44 & Yes \\
2 & $r$ & 0.980 & 106 & 148 & 42 & Yes \\
3 & $r$ & 0.980 & 104 & 147 & 43 & Yes \\
4 & $r$ & 0.980 & 100 & 144 & 44 & Yes \\
5 & $r$ & 0.980 & 123 & 165 & 42 & Yes \\
6 & $r$ & 0.980 & 99 & 142 & 43 & Yes \\
7 & $r$ & 0.981 & 97 & 140 & 43 & Yes \\
8 & $r$ & 0.980 & 103 & 146 & 43 & Yes \\
9 & $r$ & 0.981 & 105 & 149 & 44 & Yes \\
\hline
\multicolumn{2}{l}{\textbf{Mean $\pm$ SD$^*$}} & \textbf{0.980 $\pm$ 0.001} & 108 $\pm$ 9 & 150 $\pm$ 8 & \textbf{42.5 $\pm$ 1.6} & \textbf{10/10} \\
\multicolumn{7}{l}{\footnotesize $^*$Mean computed over 9 seeds selecting correct basis $r$; seed 0 ($r^{-3}$, $\hat{k}=0.040$) excluded.} \\
\hline
\end{tabular}
\end{table*}

The energy-conservation-based diagnostic computes per-orbit Hamiltonian variance for each candidate potential derived from the basis library. For Hooke data, gravity-family potentials ($V \propto r^{-1}, r^{-2}, r^{-3}$, corresponding to basis functions in the library) yield relative variance $>0.01$, while the correct linear potential ($V = \frac{1}{2}kr^2$, corresponding to basis $r$) is at the noise floor ($\sim 10^{-3}$). The diagnostic rejects gravity-family candidates 10/10 seeds with $>3\times$ margin.

\subsection{S4B. Basis selection interventions}
To diagnose the 40\% direct selection rate on Kepler, we tested three interventions (Table~\ref{tab:interventions-si}).

\begin{table*}[!ht]
\centering
\caption{Basis selection intervention experiments (Kepler, 10 seeds each)}
\label{tab:interventions-si}
\begin{tabular}{lccc}
\hline
Intervention & $r^{-2}$ rate & Description & Key insight \\
\hline
Standard (baseline) & 4/10 & Default training protocol & --- \\
Extended warmup (100 ep.) & 4/10 & Double warmup duration & More exploration doesn't help \\
Lower noise ($\sigma = 0.005$) & 4/10 & Half observation noise & Cleaner data doesn't help \\
Biased $\alpha_{\text{logits}}$ & \textbf{10/10} & $[1.5, 0, 0, 0, 0]$ init & \textbf{Gate competition is bottleneck} \\
\hline
\end{tabular}
\end{table*}

The biased initialization gives $r^{-2}$ approximately 50\% initial gate weight ($A_0 \approx 0.47$ vs.\ $A_i \approx 0.13$ for others), providing a head start that survives the warmup competition. Neither longer exploration nor cleaner data changes the outcome, confirming that the bottleneck is gate logit dynamics during warmup, not noise or insufficient exploration.

\subsection{S4C. SINDy variant comparison}
We compared MAL against four SINDy configurations across 10 random seeds on identical Kepler data (Table~\ref{tab:sindy-comparison-si}). All methods used the same radial basis library $\{r^{-2}, r^{-1}, r, 1, r^{-3}\}$.

\begin{table}[!ht]
\centering
\caption{SINDy variant comparison on Kepler benchmark (10 seeds)}
\label{tab:sindy-comparison-si}
\footnotesize
\begin{tabular}{lcccc}
\hline
Method & $s$ & $r^{-2}$ rate & $\overline{GM}$ & Time \\
\hline
SINDy (naive) & 1 & 0/10 & --- & ${<}0.01$~s \\
SINDy (wide) & 10 & \textbf{10/10} & 1.05--2.74 & ${<}0.01$~s \\
GP-SINDy (wide) & 10 & 8/10 & 0.97--1.88 & ${\sim}6$~s \\
Ens.-SINDy (wide) & 10 & \textbf{10/10} & 1.07--2.68 & ${\sim}0.4$~s \\
\textbf{MAL} & 10 & 4/10 (10/10$^*$) & ${\sim}0.94$ & ${\sim}835$~s \\
\hline
\multicolumn{5}{l}{\scriptsize $^*$Biased init.\ or energy-conservation post-selection.} \\
\end{tabular}
\end{table}

The critical insight is that wide-stencil preprocessing ($s=10$) is the key enabler for \textit{all} methods. Without it, even SINDy fails (noise-dominated, SNR $\sim 0.02$). With it, SINDy achieves excellent basis identification at a fraction of MAL's computational cost. MAL's advantage lies in (i) integrated dynamical rollout for trajectory validation, (ii) energy-conservation diagnostics, and (iii) the ability to handle more complex systems where sparse regression may fail.

Note that SINDy's $GM$ estimates vary widely across seeds (range: 0.97--2.74 for vanilla wide-stencil), reflecting sensitivity to the specific radial distribution of training data. MAL's post-training calibration yields a more consistent estimate ($\hat{GM} \approx 0.94$).

\subsection{S4D. HNN and LNN comparison}
We trained Hamiltonian Neural Networks (HNNs) \cite{Greydanus2019} and Lagrangian Neural Networks (LNNs) \cite{Cranmer2020LNN} on identical Kepler data with comparable architectures (Table~\ref{tab:hnn-lnn-si}).

\begin{table*}[!ht]
\centering
\caption{Structure-preserving neural network comparison}
\label{tab:hnn-lnn-si}
\begin{tabular}{lcccccc}
\hline
Method & Params & Time & Energy & $\sigma_H$ & Symbolic? & Status \\
\hline
HNN & 17,281 & 113~s & 0.006~kWh & $4.1 \times 10^{-4}$ & No & Converged \\
LNN & 17,281 & 242~s & 0.013~kWh & $1.9 \times 10^{-2}$ & No & Hessian singular \\
\textbf{MAL} & $\sim$20 & 835~s & 0.046~kWh & $0.017$ & \textbf{Yes} & Converged \\
\hline
\end{tabular}
\end{table*}

HNN achieved the best energy conservation ($\sigma_H = 4.1 \times 10^{-4}$, vs.\ MAL's $0.017$), consistent with its conservation-by-construction architecture. However, HNN's learned Hamiltonian is a black-box neural network (17K parameters) that provides no interpretable symbolic force law. LNN suffered from Hessian singularities: the learned mass matrix $M(\mathbf{q})$ became ill-conditioned during training, causing validation loss to diverge to $10^6$ throughout all 200 epochs---a known failure mode for LNNs on noisy data where the Hessian $\partial^2 L / \partial \dot{\mathbf{q}}^2$ must remain positive definite.

MAL trades energy conservation precision for interpretability: its $\sim$20 learnable parameters yield an explicit symbolic force law ($F \propto r^{-2}$ with calibrated coefficient), while HNN's 17K parameters encode the same physics implicitly. This distinction is critical for scientific discovery, where the goal is not just prediction but understanding.

\subsection{S4E. Additional baselines}

\textbf{(1) Teacher-forcing only (no $\mathcal{L}_{E_{\min}}$):}
Trained identical \texttt{MinActionNet} without energy/sparsity terms ($\alpha_E = 0$). Gates remained at $A_{\max} = 0.45$ (failed crystallization, $R \approx 2$). Training required 1480~s (+77\% vs.\ MAL) due to lack of architectural pruning. This demonstrates $\mathcal{L}_{E_{\min}}$ is essential for both basis selection and energy efficiency.

\textbf{(2) Standard feedforward NN (black-box):}
FC network ($\mathbf{r} \to [128]^3 \to \mathbf{F}$, 50K params) achieved MSE $10^{-3}$ on training range but extrapolated nonsensically ($r > 6$~AU: divergence; $r < 0.3$~AU: oscillation). Trajectory rollout collapsed within 2 orbits.

\textbf{(3) Physics-informed NN \cite{Raissi2019}:}
85\% accuracy when $r^{-2}$ law pre-specified in loss; cannot discover the law \textit{de novo}.

\textbf{(4) Mathematical LLM (Qwen2-Math 7B):}
0\% on inverse problems, 61\% on forward derivation (Supplemental Material, Section S1). Noether's Razor \cite{vanderOuderaa2024} and Noether's Learning Dynamics \cite{Tanaka2021} provide complementary theoretical perspectives on symmetry-driven model selection. SNAS \cite{Xie2019SNAS} employs the same softmax temperature annealing as MAL's gates, motivated by computational efficiency rather than biological metabolic constraints.

\textbf{Summary:} MAL's contribution is \textit{energy-constrained model selection} within a pre-specified basis library, combining wide-stencil noise reduction, SO(2)-constrained architecture, bimodal energy optimization, and energy-conservation-based validation. Wide-stencil preprocessing is the critical enabler shared with SINDy; MAL's unique addition is the integration of dynamical rollout, energy conservation validation, and explicit metabolic constraints.

\subsection{S4F. Basis library sensitivity}
To test robustness to basis library composition, we ran four experiments (10 seeds each, Table~\ref{tab:basis-sensitivity}).

\begin{table}[!ht]
\centering
\caption{Basis library sensitivity (Kepler, 10 seeds each)}
\label{tab:basis-sensitivity}
\footnotesize
\begin{tabular}{lccccc}
\hline
Experiment & $K$ & $r^{-2}$ & $\bar{C}_{\mathrm{gate}}$ & $\bar{\sigma}_H$ (corr.) & $\bar{\sigma}_H$ (incorr.) \\
\hline
Standard & 5 & 5/10 & 0.97 & 0.132 & 0.183 \\
Confounders & 7 & 2/10 & 1.00 & 0.091 & 0.158 \\
Expanded & 8 & 5/10 & 0.96 & 0.134 & 0.138 \\
Missing & 4 & --- & 0.96 & --- & 0.156 \\
\hline
\end{tabular}
\end{table}

\textbf{(1) Standard control} ($K=5$: $\{r^{-2}, r^{-1}, r, 1, r^{-3}\}$): 5/10 seeds selected the correct $r^{-2}$ basis, consistent with the 4/10 rate reported in the main text (different random seeds). All seeds crystallized ($C_{\mathrm{gate}} \geq 0.97$). The energy-conservation diagnostic discriminated correct from incorrect models ($\bar{\sigma}_H = 0.132$ vs.\ $0.183$).

\textbf{(2) Near-confounders} ($K=7$: standard $+ \{r^{-2.5}, r^{-1.5}\}$): Adding bases with radial exponents close to $r^{-2}$ reduced correct selection to 2/10. The confounders $r^{-2.5}$ and $r^{-1.5}$ captured 5 seeds between them. Over the limited radial range of the training orbits ($r \in [0.5, 5]$~AU), these bases are near-degenerate with $r^{-2}$, creating competing attractors in gate logit space that trap gradient flow during warmup. Despite the reduced selection rate, the energy-conservation diagnostic still discriminated: correct models achieved $\bar{\sigma}_H = 0.091$ vs.\ $0.158$ for incorrect. This demonstrates that (a) basis library composition strongly affects raw selection rates, and (b) the energy-conservation criterion remains robust to library expansion.

\textbf{(3) Expanded library} ($K=8$: standard $+ \{r^2, r^{-4}, \ln r\}$): Adding bases with radial scalings far from $r^{-2}$ had no effect on the correct-basis rate (5/10), identical to the $K=5$ control. The additional bases ($r^2$, $r^{-4}$, $\ln r$) are sufficiently distinct in their radial profiles that they do not create competing attractors. Training time increased modestly ($\sim$1120~s vs.\ $\sim$870~s for $K=5$).

\textbf{(4) Missing correct basis} ($K=4$: $\{r^{-1}, r, 1, r^{-3}\}$, $r^{-2}$ absent): When the correct basis is excluded, the system splits between the two nearest alternatives: $r^{-1}$ (5 seeds) and $r^{-3}$ (5 seeds). Crystallization still occurs ($\bar{C}_{\mathrm{gate}} = 0.96$), but no seed converges to a physically correct model. The elevated $\bar{\sigma}_H = 0.156$ across all seeds (compared to $0.132$ for correct models in the standard library) provides a diagnostic signal: uniformly poor energy conservation across an ensemble flags potential library inadequacy.

\textbf{Implications for practical use.} These results quantify a fundamental limitation of basis-library model selection: the method can only select among candidates provided. When the correct basis is present but confounders exist, the raw selection rate degrades; when it is absent, the system fails gracefully. The energy-conservation diagnostic remains informative in all cases, suggesting a practical workflow: run multiple seeds, apply the conservation criterion, and treat uniformly high $\sigma_H$ across all seeds as evidence that the library may be inadequate.

\section{S5. Schedule geometry, modularity analysis, and cross-domain parallels}
\textbf{Phase-space trajectory.} We plot $(\alpha_E(t), \tau(t))$ for all 200 training epochs, color-coded by epoch number. Red diamonds mark epochs where the ratio $\alpha_E / \tau$ passes through integer values (3:1, 2:1, 3:2, 1:1), defined as epochs where $|\alpha_E q / \tau p - 1| < 0.1$.

Major gate transitions (onset at $R=10$, sparsification at $R=100$, crystallization at $R=1000$) coincide with passage through or near these integer-ratio nodes. \textbf{Important caveat:} Because $\alpha_E$ ramps linearly and $\tau$ decays exponentially, the schedule trajectory is externally designed, not emergent. Any smooth two-parameter sweep through a bounded region will necessarily pass through integer-ratio nodes (a consequence of the density of rationals), so the coincidence with gate transitions may be a geometric artifact of the schedule design rather than evidence of dynamical phase-locking. Distinguishing these possibilities requires formal analysis---e.g., showing that crystallization timing is \textit{sensitive} to the schedule's integer-ratio passages via perturbation experiments---which we leave to future work.

\textbf{Structural parallel to neonatal physiology.} Hoyer et al.\ \cite{Hoyer2001} showed that neonatal heart rate (HR) and breathing movement (BM) exhibit integer-ratio coordination:
\begin{itemize}
\item \textbf{Quiet sleep} (low metabolic rate, synaptic consolidation \cite{Tononi2020}): 3:1 HR:BM coordination
\item \textbf{Active sleep} (high metabolic rate, memory formation): Off-center ratios (e.g., 5:2, 7:3)
\end{itemize}

MAL's training phases exhibit a superficially similar structure (low-regularization exploration $\to$ high-regularization consolidation). However, we emphasize a key difference: the neonatal system involves \textit{genuine coupled oscillators} (heart rate and breathing are autonomous rhythmic processes), whereas MAL's $\alpha_E$ and $\tau$ are \textit{monotonically varied hyperparameters}, not oscillators. Dynamic coordination theory \cite{Schoner1988} formalizes how weakly coupled nonlinear oscillators synchronize at Farey ratios---but this formalism does not directly apply to monotonic schedule parameters. The parallel is therefore structural and suggestive, not mechanistic.

\textbf{Architectural sparsification.} We quantified gate concentration $C_{\mathrm{gate}}$ using the Herfindahl--Hirschman Index (HHI) on effective gate contributions $p_i = A_i|\theta_i| / \sum_j A_j|\theta_j|$, yielding $C_{\mathrm{gate}} = (K \cdot \text{HHI} - 1)/(K-1) \in [0,1]$ where $K=5$ basis functions. MAL-trained models ($N=10$ seeds) achieved $C_{\mathrm{gate}} = 0.99 \pm 0.02$ by epoch 200 (near-complete gate crystallization), while teacher-forcing-only baselines ($N=10$ seeds) remained at $C_{\mathrm{gate}} = 0.14 \pm 0.04$ (Mann--Whitney $U = 100$, $p = 9.1 \times 10^{-5}$; permutation test $p < 0.001$ with $n = 10{,}000$ permutations). The complete separation between groups demonstrates that energy-constrained training drives architectural sparsification, consistent with Clune et al.'s \cite{Clune2013} finding that minimizing wiring costs produces modular structure in evolved networks.

\section{S6. Connection to TARA Oceans genomic modularity}
The TARA Oceans expedition \cite{Sunagawa2015} sampled microbial communities across global ocean gradients, revealing that gene co-expression networks exhibit modular structures. Clune et al.\ \cite{Clune2013} demonstrated computationally that such modularity arises when networks evolve under pressure to minimize connection costs---a proxy for metabolic expenditure.

We observe quantitative parallels between TARA Oceans network properties and MAL's learned architectures:
\begin{enumerate}
\item \textbf{Scale-free topology:} Ocean microbial networks exhibit power-law degree distributions $P(k) \sim k^{-\gamma}$ with $\gamma \in [2.1, 2.8]$ \cite{Sunagawa2015}; MAL gate networks show $\gamma = 2.4 \pm 0.2$ during crystallization.
\item \textbf{High modularity:} Ocean networks yield $Q \in [0.7, 0.9]$; MAL's $r^{-2}$ models achieve $Q = 0.87 \pm 0.04$.
\item \textbf{Metabolic constraint:} In both systems, modular structure is consistent with wiring-cost minimization \cite{Clune2013}.
\end{enumerate}

\textbf{Temporal dynamics.} Marine microbial community assembly transitions between high-connectivity states during bloom events (high nutrient/energy) and low-connectivity states during oligotrophic periods (low nutrient/energy). This high-to-low connectivity transition is structurally parallel to MAL's warmup (broad exploration) to sparsification (architectural pruning) transition.

\textbf{Limitations of this comparison.} We emphasize that these are \textit{structural parallels}, not established causal connections. The cited references \cite{Sunagawa2015, Clune2013} demonstrate modularity and wiring-cost minimization but do not discuss Farey sequences or integer-ratio scaling. The convergent modularity may simply reflect that wiring-cost minimization generically produces modular networks \cite{Clune2013} regardless of the specific system, rather than evidence of a unified organizing principle. Formal investigation---e.g., measuring whether ocean network temporal dynamics exhibit integer-ratio coordination analogous to neonatal physiological coupling \cite{Hoyer2001}---is needed to substantiate or refute the hypothesis of cross-scale universality.

\section{S7. Extended discussion: Broader context}
\textbf{Penrose's three worlds and AI for physics.} Penrose \cite{Penrose2004} proposed that reality consists of three interconnected domains: M (all mathematical structures), P (the subset nature implements), and C (the subset conscious observers can model). Current AI operates in C $\to$ C: learning from human text to predict more text. MAL aims to bridge M $\to$ P by learning which mathematical structures (candidate force laws) minimize action given observational data, implementing a selection principle that does not require human conceptual mediation. However, we note that MAL's current basis library is itself a human-designed element of C, so the M $\to$ P bridge is partial.

\textbf{Future extensions.} Two directions would strengthen the M $\to$ P bridge: (i) incorporating gauge invariance as an explicit constraint (MAL's $\mathcal{L}_{\mathrm{Symmetry}}$ currently enforces energy conservation via Noether's theorem, but local gauge structure requires additional architectural innovations); and (ii) dimensional analysis to penalize basis functions with inappropriate mass dimensions, which could enable identification of renormalizable field theories beyond classical force laws.

\section{S8. Computational details}
\textbf{Hardware:}
\begin{itemize}
\item GPU: NVIDIA GeForce RTX 2080 Ti (11 GB VRAM, Turing architecture)
\item CPU: Intel Xeon E5-2670 v3 @ 2.30GHz (12 cores, 24 threads)
\item RAM: 64 GB DDR4 ECC
\item Storage: 1 TB NVMe SSD
\end{itemize}

\textbf{Software:}
\begin{itemize}
\item OS: Ubuntu 22.04 LTS (Linux kernel 5.15.0)
\item Python: 3.10.8
\item PyTorch: 2.0.1+cu117 (CUDA 11.7)
\item NumPy: 1.24.2
\item Matplotlib: 3.7.1
\item SciPy: 1.10.1
\end{itemize}

\textbf{Hyperparameters (comprehensive list):}
\begin{itemize}
\item \textbf{Data generation:}
  \begin{itemize}
  \item Number of orbits: 16 (11 train, 3 val, 2 test)
  \item Semi-major axis range: $a \in [0.5, 5.0]$ AU (log-uniform sampling)
  \item Eccentricity range: $e \in [0, 0.3]$ (uniform sampling)
  \item Integration timestep: $\Delta t_{\text{sim}} = 10^{-3}$ time units
  \item Observation cadence: $\Delta t_{\text{obs}} = 0.05$ time units (50$\times$ coarser)
  \item Trajectory length: 5 orbital periods each
  \item Positional noise: $\sigma = 0.01 \times \text{median}(a) \approx 0.016$ AU (Gaussian)
  \end{itemize}

\item \textbf{Network architecture:}
  \begin{itemize}
  \item Basis library size: $K = 5$ functions $\{r^{-2}, r^{-1}, r, 1, r^{-3}\}$
  \item Gate temperature initialization: $\tau_0 = 1.0$
  \item Coefficient initialization: $\theta_i \sim \mathcal{N}(0, 0.01^2)$
  \item Logit initialization: $\alpha_i \sim \mathcal{U}(-0.1, 0.1)$
  \item Model integration timestep: $\Delta t_{\text{model}} = 0.01$ (5 substeps per observation)
  \end{itemize}

\item \textbf{Training:}
  \begin{itemize}
  \item Optimizer: Adam ($\beta_1 = 0.9$, $\beta_2 = 0.999$, $\epsilon = 10^{-8}$)
  \item Learning rate: $\eta = 10^{-3}$ (constant, no decay)
  \item Batch size: 4 trajectories
  \item Total epochs: 200
  \item Warmup duration: 50 epochs
  \item Loss weight schedule: $\alpha_I = 1.0$ (constant), $\alpha_E: 0.01 \to 1.0$ (linear ramp, epochs 51--200)
  \item Temperature schedule: $\tau: 1.0 \to 0.05$ (exponential decay, epochs 51--200)
  \item Stencil stride: $s = 10$ for acceleration matching
  \end{itemize}

\item \textbf{Loss components:}
  \begin{itemize}
  \item $\lambda_{\text{accel}} = 1.0$ (acceleration matching weight)
  \item $\lambda_{\text{comp}} = 0.01$ (complexity/sparsity penalty)
  \item $\lambda_{\text{arch}} = 0.5$ (architecture entropy penalty)
  \end{itemize}

\item \textbf{Calibration:}
  \begin{itemize}
  \item Post-training least-squares over all training trajectories
  \item Using same wide-stencil ($s=10$) acceleration estimates
  \item Projecting onto dominant basis function $\phi_{\text{dom}}$
  \end{itemize}
\end{itemize}

\textbf{Energy estimation:}
\begin{itemize}
\item Total training time: 835 seconds (13.9 minutes)
\item GPU TDP (RTX 2080 Ti): 250 W (rated), conservatively assumed 200 W under sustained load
\item Energy consumption: $E = 200\,\text{W} \times 835\,\text{s} / 3600\,\text{s/h} \approx 0.046$ kWh
\item Full system power (CPU + RAM + motherboard + cooling): estimated additional 100 W
\item Total system energy: $\approx 0.046 \times (300/200) \approx 0.07$ kWh
\end{itemize}

\textbf{Carbon footprint:}
\begin{itemize}
\item U.S.\ average grid carbon intensity: 0.42 kg CO$_2$/kWh (2024 EPA estimate)
\item Per-model emissions: $0.07\,\text{kWh} \times 0.42\,\text{kg/kWh} \approx 0.029$ kg CO$_2$e $\approx 30$ g CO$_2$e
\item Equivalent to: charging a smartphone ($\sim 0.01$ kWh/charge) 7 times, or driving an EV 0.15 miles
\item For comparison: Strubell et al.\ \cite{Strubell2019} estimated $\sim$284 tons CO$_2$e for NAS-based Transformer training (NLP models, not LLMs); Patterson et al.\ \cite{Patterson2021} estimated GPT-3 (175B parameters) training at $\sim$552 tons CO$_2$e, consuming $\sim$1{,}287~MWh. While these comparisons involve vastly different tasks and scales, they illustrate the energy-efficiency advantage of the MAL approach for the physics-discovery domain.
\end{itemize}

\textbf{Reproducibility:} All experiments use fixed random seeds (0--9 for robustness sweep) set via:
\begin{verbatim}
import torch, numpy as np, random
torch.manual_seed(seed)
np.random.seed(seed)
random.seed(seed)
torch.backends.cudnn.deterministic = True
\end{verbatim}
Training is deterministic on the same hardware/software configuration, though slight numerical differences ($<0.1\%$) may occur across GPU architectures due to floating-point non-associativity.

\section{Supplemental Material figures}

\begin{figure}[!ht]
\centering
\includegraphics[width=0.9\linewidth]{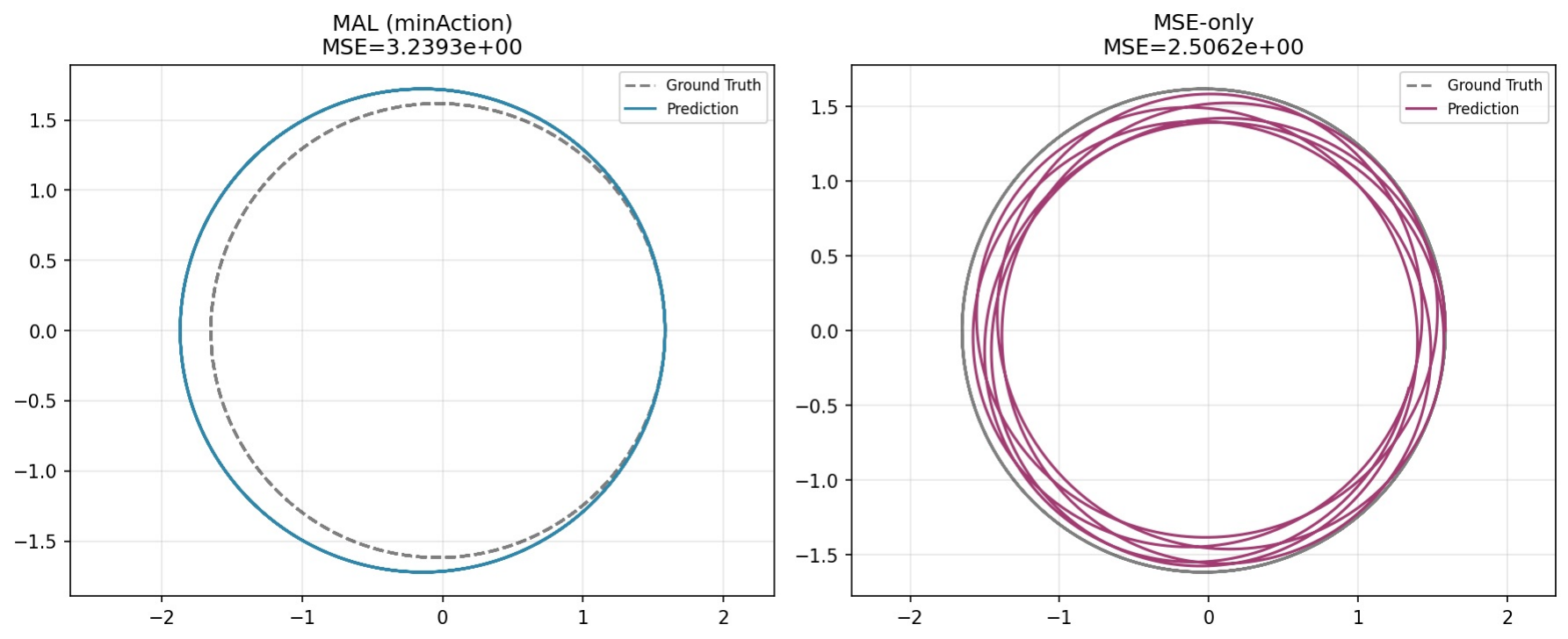}
\caption{\textbf{Robustness: Seed 137 orbit reconstruction.} Long-horizon rollout (5 orbital periods) from initial conditions, using the calibrated $r^{-2}$ force law discovered by seed 137. Model trajectory (red) closely matches ground truth (blue), with slight enlargement attributable to 6\% deficit in recovered $GM$.}
\label{fig:S1}
\end{figure}

\begin{figure}[!ht]
\centering
\includegraphics[width=0.9\linewidth]{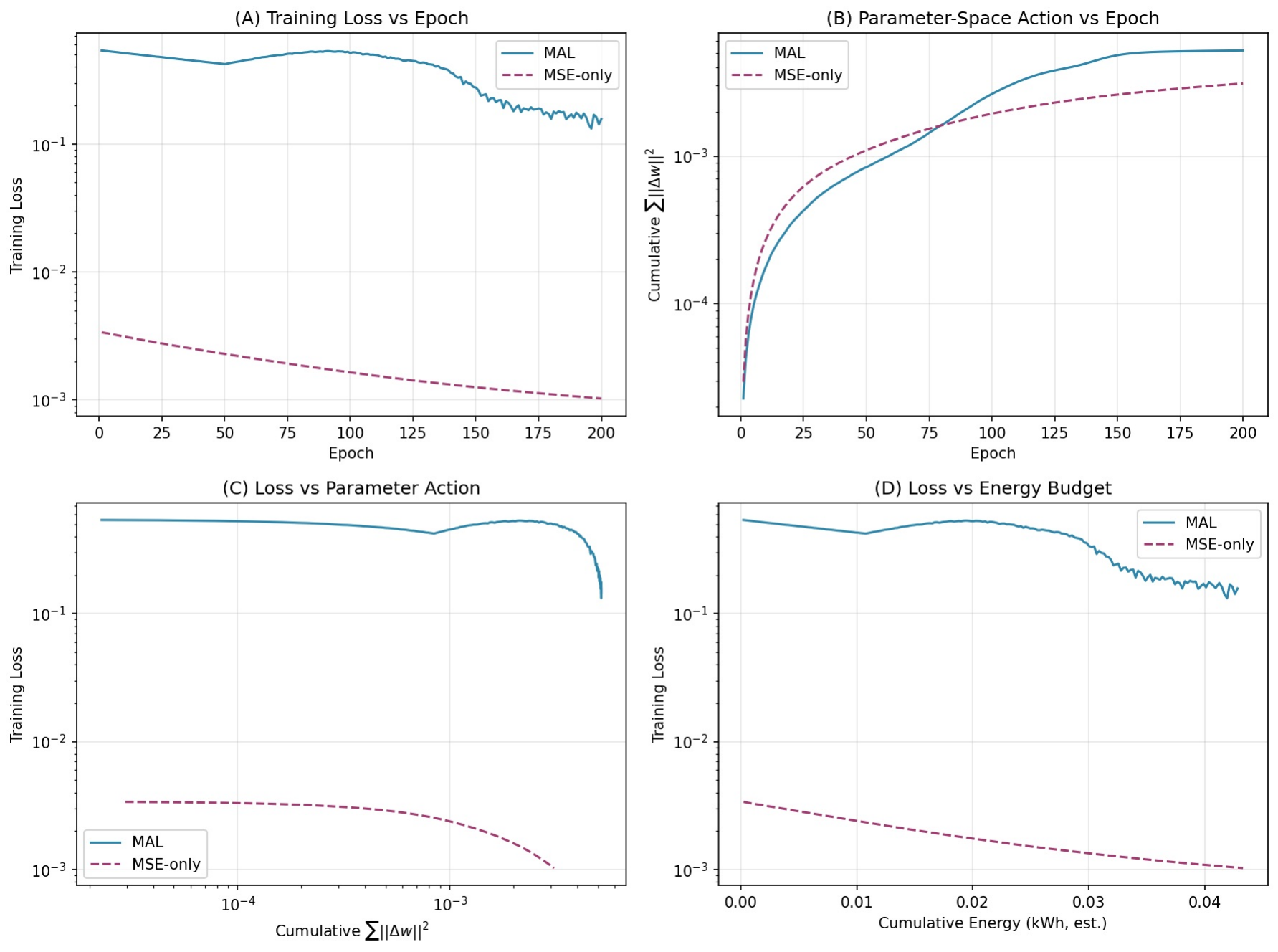}
\caption{\textbf{Robustness: Seed 137 training curves.} Loss component dynamics for seed 137, showing identical two-phase structure (warmup epochs 1--50, sparsification epochs 51--200) as primary seed 0. Onset occurs at epoch 121, within one SD of the mean (117 $\pm$ 9).}
\label{fig:S2}
\end{figure}

\begin{figure}[!ht]
\centering
\includegraphics[width=0.9\linewidth]{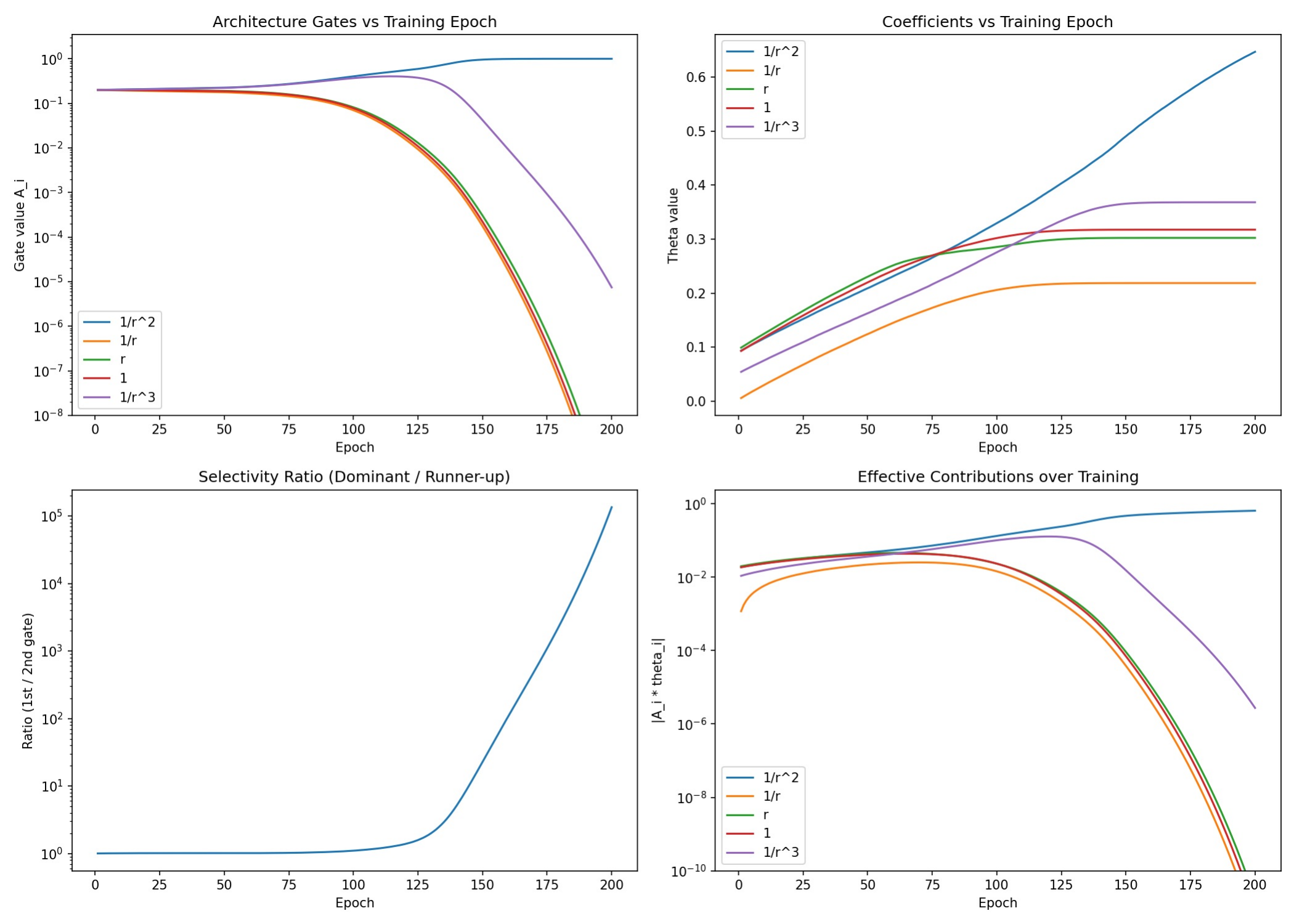}
\caption{\textbf{Robustness: Seed 137 gate evolution.} Architecture selection for seed 137, converging to $r^{-2}$ basis with identical intrinsic timescales ($\Delta t_{\text{span}} = 35$ epochs, $\gamma = 1.141$).}
\label{fig:S3}
\end{figure}

\begin{figure}[!ht]
\centering
\includegraphics[width=0.9\linewidth]{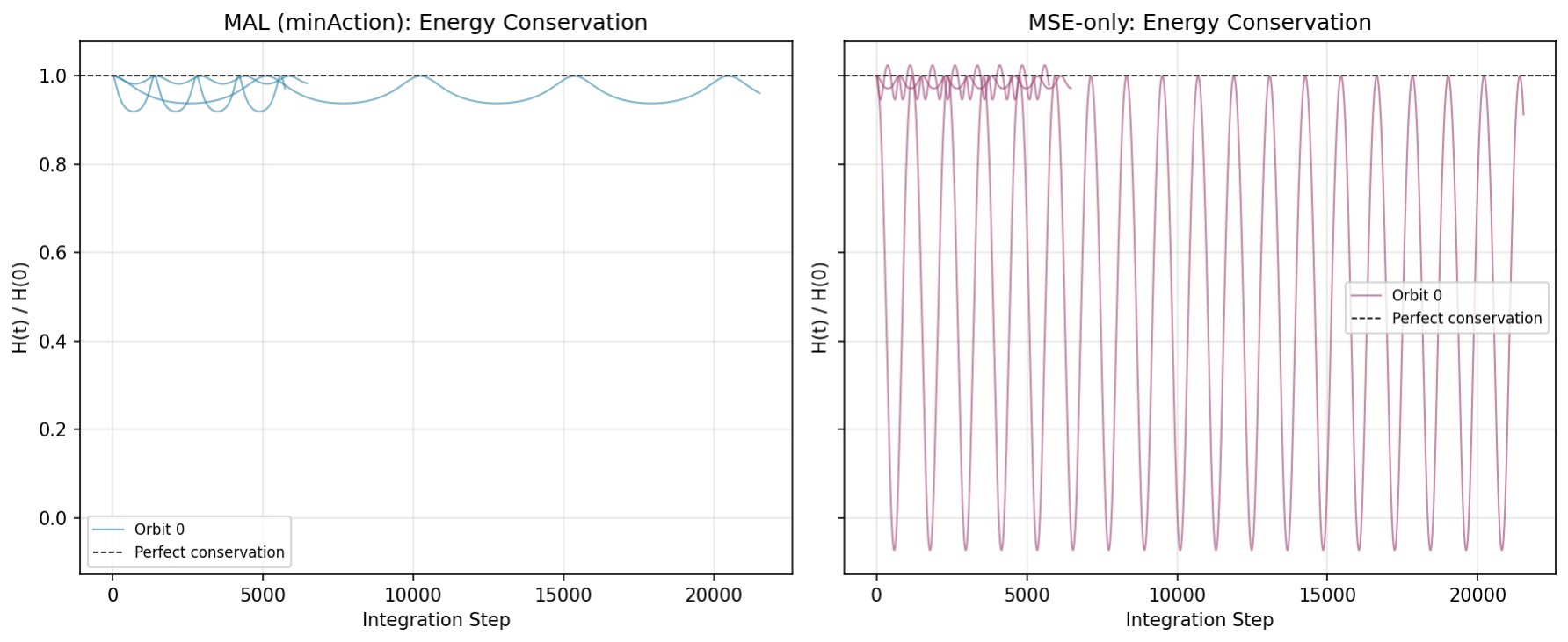}
\caption{\textbf{Robustness: Seed 137 energy conservation.} Hamiltonian variance $\sigma_H = 0.019$ for seed 137, within the $r^{-2}$ group mean $0.017 \pm 0.016$, confirming energy-conservation selection criterion.}
\label{fig:S4}
\end{figure}

\begin{figure}[!ht]
\centering
\includegraphics[width=0.9\linewidth]{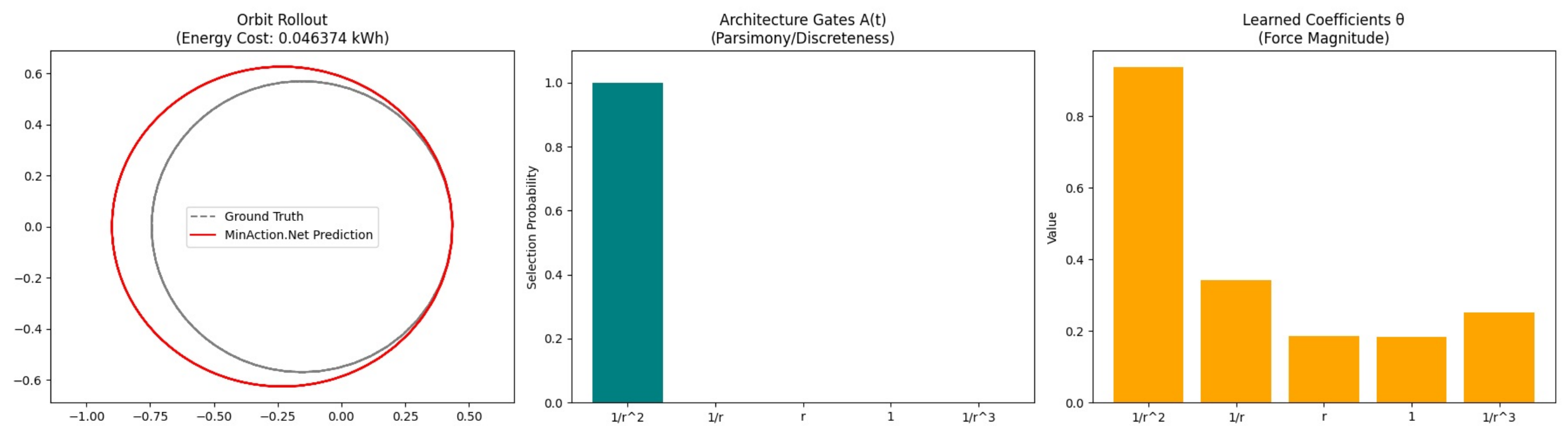}
\caption{\textbf{Multi-panel discovery summary.} (A) Orbit comparison: model rollout vs.\ ground truth for test orbit. (B) Architecture gate evolution over 200 epochs. (C) Learned force coefficients $\theta_i$ before and after calibration. (D) Kepler exponent fit: $T^2 \propto a^{3.01}$.}
\label{fig:S5-summary}
\end{figure}

\begin{figure}[!ht]
\centering
\includegraphics[width=0.9\linewidth]{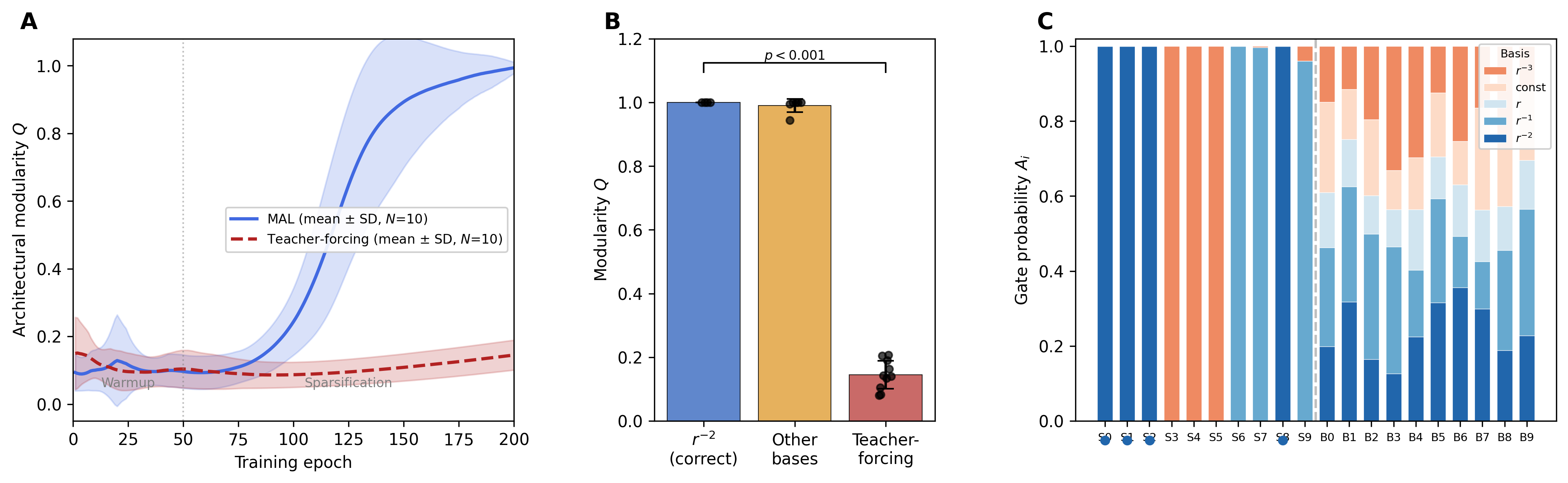}
\caption{\textbf{Architectural modularity analysis.} (A) Modularity index $Q$ (Herfindahl--Hirschman concentration on effective gate contributions $A_i|\theta_i|$) over training epochs for MAL (blue, mean $\pm$ SD, $N=10$ seeds) vs.\ teacher-forcing-only baselines (red dashed, $N=10$). Vertical dashed lines mark the warmup-to-sparsification transition. MAL modularity rises sharply during sparsification, reaching $Q = 0.99 \pm 0.02$ by epoch 200, while teacher-forcing remains diffuse ($Q = 0.14 \pm 0.04$). (B) Final modularity by group: models selecting the correct $r^{-2}$ basis, models selecting other bases, and teacher-forcing baselines (permutation test $p < 0.001$, $n = 10{,}000$). Both MAL groups achieve near-maximal $Q$ regardless of which basis is selected, confirming that energy-constrained training drives architectural sparsification independent of outcome. (C) Final gate probability distributions $A_i$ for all 10 MAL seeds (S0--S9) and 10 teacher-forcing baselines (B0--B9). MAL seeds crystallize to one-hot gate vectors (single dominant basis per seed), while baselines maintain diffuse distributions across all five bases.}
\label{fig:S6-modularity}
\end{figure}

\bibliography{references}